\documentclass[10pt,journal,compsoc]{IEEEtran}
\ifCLASSOPTIONcompsoc
\usepackage[nocompress]{cite}
\else
\usepackage{cite}
\fi
\ifCLASSINFOpdf
 \usepackage[pdftex]{graphicx}
\else
\usepackage[dvips]{graphicx}
\DeclareGraphicsExtensions{.eps}
\fi
\usepackage{amsmath}
\usepackage{amssymb}
\usepackage{amsthm}

\usepackage{array}
\usepackage{hyperref}

\usepackage[linesnumbered,ruled]{algorithm2e}

\theoremstyle{definition}

\let\oldnl\nl
\newcommand{\nonl}{\renewcommand{\nl}{\let\nl\oldnl}}

\renewcommand\thesection{\Alph{section}}

\usepackage{multirow}
\usepackage[font=scriptsize]{subfig}
\usepackage{epstopdf}
\usepackage{subfiles}
\usepackage{soul}
\usepackage{color, xcolor}
\usepackage{verbatim}

\soulregister\cite7 
\soulregister\eqref7
\begin{document}
	%
	\title{Adversarial Reinforced Instruction Attacker for Robust Vision-Language Navigation}
	%
	%
	%
	%
	
	\author{Bingqian~Lin, Yi Zhu, Yanxin Long, Xiaodan Liang\IEEEauthorrefmark{2}, Qixiang Ye,  Liang Lin
	\IEEEcompsocitemizethanks{
	\IEEEcompsocthanksitem 
	\IEEEauthorrefmark{2}Xiaodan Liang is the corresponding author.\protect\\	\IEEEcompsocthanksitem Bingqian Lin, Yanxin Long and Xiaodan Liang are with Shenzhen Campus of Sun Yat-sen University, Shenzhen, China. \protect\\
			E-mail:\{linbq6@mail2.sysu.edu.cn,longyx9@mail2.sysu.edu.cn, liangxd9@mail.sysu.edu.cn\}
		\IEEEcompsocthanksitem  Liang Lin is with Sun Yat-sen University, Guangzhou, China.\protect\\
			E-mail:\{linliang@ieee.org\}
			\IEEEcompsocthanksitem Yi Zhu, Qixiang Ye are with University of Chinese
	Academy of Sciences (UCAS), Beijing, China.
	\protect\\
	E-mail: \{zhu.yee@outlook.com, qxye@ucas.ac.cn\}
	
	}
		}
	
	%
	%

	\markboth{IEEE TRANSACTIONS ON PATTERN ANALYSIS AND MACHINE INTELLIGENCE}%
	{Shell \MakeLowercase{\textit{et al.}}: Bare Demo of IEEEtran.cls for Computer Society Journals}
	%



	
	\IEEEtitleabstractindextext{%
		\begin{abstract}
		Language instruction plays an essential role in the natural language grounded navigation tasks. However, navigators trained with limited human-annotated instructions may have difficulties in accurately capturing key information from the complicated instruction at different timesteps, leading to poor navigation performance. In this paper, we exploit to train a more robust navigator which is capable of dynamically extracting crucial factors from the long instruction, by using an adversarial attacking paradigm. Specifically, we propose a Dynamic Reinforced Instruction Attacker (DR-Attacker), which learns to mislead the navigator to move to the wrong target by destroying the most instructive information in instructions at different timesteps. By formulating the perturbation generation as a Markov Decision Process, DR-Attacker is optimized by the reinforcement learning algorithm to generate perturbed instructions sequentially during the navigation, according to a learnable attack score. Then, the perturbed instructions, which serve as hard samples, are used for improving the robustness of the navigator with an effective adversarial training strategy and an auxiliary self-supervised reasoning task. Experimental results on both Vision-and-Language Navigation (VLN) and Navigation from Dialog History (NDH) tasks show the superiority of our proposed method over state-of-the-art methods. Moreover, the visualization analysis shows the effectiveness of the proposed DR-Attacker, which can successfully attack crucial information in the instructions at different timesteps. Code is available at \url{https://github.com/expectorlin/DR-Attacker}.
		\end{abstract}
		
		}

	\maketitle

	\IEEEdisplaynontitleabstractindextext

	%
	\IEEEpeerreviewmaketitle



	%
		
\IEEEraisesectionheading{\section{Introduction}\label{sec:introduction}}

	%
	%
	%
	%

\IEEEPARstart{N}{atural} language grounded visual navigation task asks an embodied agent to navigate to a goal position following language instructions \cite{anderson2018vision,thomason2019vision,qi2020reverie,nguyen2019help,chen2019touchdown}. It has raised widely research interests  in recent years since an instruction-following navigation agent is more flexible and practical in many real-world applications, such as personal assistants and in-home robots.
To accomplish successful navigation, the agent needs to extract the key information, e.g., visual objects, specific rooms or navigation directions, from the long instruction according to dynamic visual observation for guiding navigation at each timestep.
However, due to the complexity and semantic ambiguity of the natural language, it is hard for the navigators to effectively learn cross-modality alignment and capture accurate semantic intentions from the instruction by training with limited human-annotated instruction-path data.
	
Prior works mainly employed the data augmentation strategy to solve the data scarcity in  navigation tasks \cite{fried2018speaker,tan2019learning,fu2020counterfactual}.  
\cite{fried2018speaker} proposed a speaker-follower framework to generate augmented instructions within randomly sampled paths. 
However, generating a  large amount of the whole instructions is at high costs and may not contribute to the emphasis of the most instructive  information. 
 \cite{tan2019learning} and \cite{fu2020counterfactual} put more focus on creating challenging augmented paths and diverse visual scenes,  while generated augmented instructions by employing the speaker model in \cite{fried2018speaker} directly. Therefore, the enhancement of the instruction understanding ability of the navigator might also be limited.

\begin{figure*}
	\centering
	\includegraphics[width=1.0\linewidth]{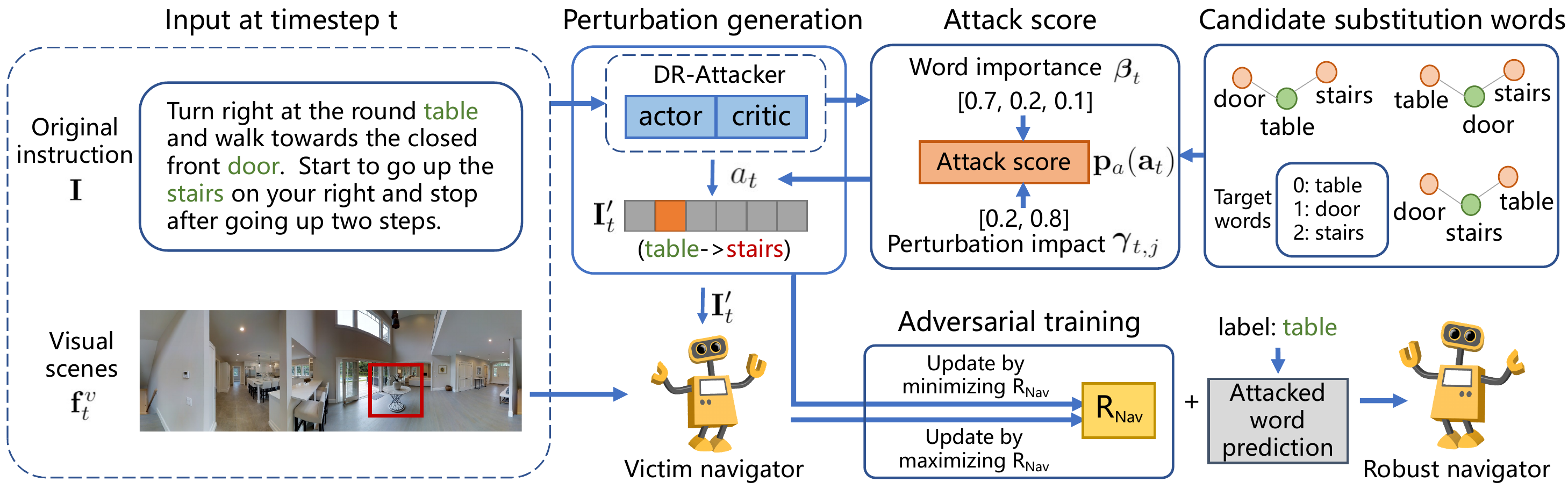}
	\caption{
		The overview of our proposed method. At  timestep $t$, the DR-Attacker receives the visual observation and original instruction, and generates the perturbed instruction $\mathbf{I}_{t}'$ by substituting the selected  target word with the best candidate word according to the attack score. The victim navigator, which receives the perturbed instruction, is enforced to maximize the navigation reward $R_{Nav}$ with an adversarial  setting and reasoning the actual attacked words by DR-Attacker to enhance the model robustness.
	}
	\vspace{-0.6cm}
	\label{fig:framework}
	
\end{figure*}

In recent years, there have been increasing attentions in designing the adversarial attacks for natural language processing (NLP) tasks to verify and improve the robustness of NLP models \cite{araujo2020on,li2020bert,ebrahimi2018on,ren2019generating}. 
Inspired by this, we consider the following question: Can we design adversarial attacks on the instruction to generate helpful adversarial samples  for improving the robustness of the navigator?
A simple way to generate adversarial  instructions is to  borrow the existing attack methods on NLP \cite{ren2019generating,zang2020word} tasks directly.
However, it is difficult since existing adversarial attacks on NLP are often optimized by some classification-based goal functions  \cite{araujo2020on,ren2019generating}, which are unreachable in the navigation tasks. 
Moreover, the key  instruction information for navigation changes dynamically while these attack methods developed on NLP are designed in the static setting. 

In this paper, we make the first attempt for introducing the adversarial attacks on the language  instruction of  navigation tasks to improve the robustness of navigators.
Specifically, we propose a Dynamic Reinforced Instruction Attacker (DR-Attacker), which learns to minimize the navigation reward by dynamically {\it destroying} key instruction information and generating  perturbed instruction  at each timestep. 
Then, an effective adversarial training strategy is adopted to improve the robustness of the navigator, by  asking it to maximize  the navigation reward with the perturbed instruction. 
To encourage the agent to be aware of actual key information  and improve the fault-tolerance ability with perturbed instruction, an auxiliary self-supervised  reasoning task is also introduced for the  navigator, requiring it to distinguish the actual attacked word of the DR-Attacker at each timestep according to the instruction and current visual observation. As a result, more accurately the DR-Attacker attacks the important instruction information, more possible that the agent is able to  capture the actual key information for navigation.

	
Since navigation is a sequential decision making problem without direct classification-based objectives, we formulate the perturbation generation as a Markov Decision Process, and present a reinforcement learning (RL) resolution to generate the perturbed instructions by misleading the navigator to move to the wrong target position.
At each timestep, the policy agent, i.e., our proposed DR-Attacker,   substitutes the most crucial target word in the current instruction with the best candidate substitution word which has maximum perturbation impact,  according to a learnable attack score. 
As a result, the DR-Attacker can learn to highlight the important parts in instructions to generate  adversarial samples at different timesteps.
To enhance the  navigation robustness, the victim navigator, which receives the perturbed instruction, is enforced to be immune to the perturbation under the adversarial setting, as well as correctly reasoning the actual attacked words by the DR-Attacker. 
The overview of our proposed method is presented in Figure \ref{fig:framework}. 
Suppose a person receives the perturbed instruction $\mathbf{I}_{t}'$ where the word ``table'' is substituted with the word ``stairs''. With the good understanding of the instruction and visual environment, he can distinguish the noisy word and still make the correct navigation decision.
Therefore, the perturbed instructions, which can be viewed as hard negative samples, can effectively encourage the victim navigator to understand the  multi-modality observations and have the self-correction ability thus become more robust.

Experimental results on both Navigation from Dialog History (NDH) and Vision-and-Language Navigation (VLN) 
show the superiority of the proposed method over other competitors. Moreover, the quantitative and qualitative results show the effectiveness of the proposed DR-Attacker, which causes significant navigation performance drop by only  disturbing most  crucial instruction information.

The merits of our proposed DR-Attacker are summarized as follows:
First, DR-Attacker can generate perturbed instruction dynamically by capturing and destroying key  instruction information in different navigation timesteps.
Second, DR-Attacker can be optimized via gradient-based methods under the unsupervised setting, by formulating the perturbation generation as a sequential decision making problem. 
Last but not least, the adversarial samples produced by DR-Attacker are beneficial for improving the model robustness.

The main contributions of this paper are summarized as follows:
\begin{itemize}
    \item 
    We take the first step to introduce the adversarial attack on the language instruction of navigation tasks to learn  robust navigators.
    Different from existing adversarial attacking paradigm developed on NLP tasks which are generally static, the proposed adversarial attack is dynamic during the navigation process. 
    \item By formulating the perturbation generation as a Markov Decision Process, the proposed instruction attacker, called Dynamic Reinforced Instruction Attacker (DR-Attacker), can be optimized by the reinforcement learning algorithm to achieve effective perturbation, without the need of classification-based objectives. 
    \item To improve the robustness of the navigator, an alternative adversarial training strategy and an auxiliary self-supervised reasoning task are employed to train the navigator on perturbed instructions, which can effectively enhance the cross-modal understanding ability of the navigator.
    \item Experimental results on two popular natural language grounded visual navigation tasks, i.e., Vision-and-Language Navigation (VLN) and Navigation from Dialog History (NDH) show that the model robustness can be effectively enhanced by the proposed method. Moreover, both the quantitative results and visualized results show the effectiveness of the proposed DR-Attacker. 
\end{itemize}

The remainder of this paper is organized as follows. Section \ref{related work} gives a brief review of the related work. Section \ref{method} describes the problem setup of natural language grounded visual navigation tasks and then introduces our proposed method. Experimental results are provided in Section  \ref{experiment}. Section \ref{conclusion} concludes the paper and presents some outlook for future work.

	\section{Related Work}
    \label{related work}

	\subsection{Natural Language Grounded Visual Navigation}
Natural language grounded visual navigation tasks \cite{wang2020environment,qi2020reverie,nguyen2019help,thomason2019vision,anderson2018vision,chen2019touchdown,nguyen2019vision} have attracted extensive research interests in recent years since they are practical and pose great challenges for vision-language understanding tasks \cite{antol2015vqa,vries2017guesswhat,das2017visual,you2016image}. 
In this paper, we  mainly focus on two natural language grounded navigation tasks, namely, Vision-and-Language Navigation (VLN) \cite{anderson2018vision} and Navigation from Dialog History (NDH) \cite{thomason2019vision}.

\textbf{Vision-and-Language Navigation (VLN)} \cite{anderson2018vision,fried2018speaker,tan2019learning,fu2020counterfactual} was first proposed by \cite{anderson2018vision}, where a navigation agent is asked to move to the goal position following the navigation instruction.  Specifically, the instruction is a sequence of declarative sentences such as ``Walk down stairs. Walk past the chartreuse ottoman in the TV room. Wait in the bathroom door threshold.'' Therefore, to successfully navigate to the goal position, the agent needs to understand the instruction well and learn to ground the instruction to visual observations. To achieve this, \cite{wang2019reinforced} proposed Reinforced Cross-Modal Matching (RCM) approach to enforce cross-modal grounding both locally and globally via reinforcement learning (RL). \cite{ma2019self} designed visual-textual co-grounding module to distinguish different instruction parts as the ones have completed and the ones need to complete regarding visual observations. To better encourage the navigator to sufficiently understand the diverse instructions and navigation environments,   existing works  adopted the data augmentation strategy \cite{tan2019learning,fried2018speaker,fu2020counterfactual} to solve the data scarcity in the original dataset. A speaker-follower model was proposed by \cite{fried2018speaker} to produce augmented instructions with randomly-sampled paths.  \cite{tan2019learning} proposed Environmental Dropout to create new (environment, path, instruction) triplets while utilizing the speaker model in \cite{fried2018speaker} directly for generating the augmented instructions. 

The Cooperative Vision-and-Dialog Navigation (CVDN) dataset was recently proposed by \cite{thomason2019vision} and \textbf{Navigation from Dialog history (NDH)} is a task proposed on CVDN dataset, which requires an agent to move towards the goal position following a sequence of dialog history. Although the visual scenes in CVDN dataset are similar to the R2R dataset proposed on VLN task \cite{anderson2018vision}, the instruction in the CVDN dataset, which is composed of dialog history and current question-answer pair, is harder for the agent to understand and perform visual grounding since it is longer and more complicated than the instruction on VLN task.  
To better explore useful textual information for successful navigation, \cite{zhu2020vision} proposed Cross-modal Memory Network (CMN) to exploit the rich information in dialog history. 
\cite{hao2020towards} employed a pretraining scheme by using image-text-action triplets for improving the instruction understanding and cross-modality alignment. 

While existing methods have achieved some improvements in enhancing the instruction understanding by data augmentation \cite{fried2018speaker,tan2019learning,fu2020counterfactual} or pretraining \cite{hao2020towards,li2019robust}, the quality of the augmented instructions is rarely noticed, leading to limited  improvement of the model robustness. In contrast, we adopt an adversarial attack paradigm to encourage the generation of meaningful adversarial instructions, which can serve as hard augmented samples to better enhance the navigation robustness. 

	\subsection{Adversarial Attacks in NLP}

	
Adversarial attacks have been widely used in the image domain \cite{cemgil2020adversarially,zhang2019defense,salman2019provably,feng2019learning,mopuri2019generalizable} to validate the robustness of the deep neural network models \cite{he2016deep,simonyan2015very}. In recent years, many researchers of NLP fields put their focus on introducing adversarial attacks for the NLP tasks, which can serve as a powerful tool for evaluating the model vulnerability, and more importantly, improving the robustness of NLP models   \cite{wang2019improving,ren2019generating,zhu2019freelb,cheng2019robust,jones2020robust}. The key principle of adversarial attacks is to impose imperceptible perturbation by human on the original input while easily fool the neural model to make the incorrect prediction. Most adversarial attacks on NLP tasks are word-level attacks \cite{zang2020word,ren2019generating} or character-level attacks \cite{araujo2020on,ebrahimi2018on}. HotFlip \cite{ebrahimi2018hotflip} introduced white-box adversarial samples based on an atomic flip operation to trick a character-level neural classifier. \cite{zang2020word} proposed a word-level attack model based on sememe-based word substitution method and particle swarm optimization-based search algorithm, which was implemented on Bi-LSTM \cite{conneau2017supervised} and BERT \cite{devlin2019bert}. 
Due to the discrete characteristic of the natural language, the imposed adversarial attacks on the language, such as inserting, removing or replacing a specific character or word, can easily change the meaning or break the grammaticality and naturality of the original sentence \cite{zang2020word,wang2021infobert}. Therefore, the adversarial perturbation on the language is essentially easy to be perceived by human rather than that in image.

Our introduced attack on the instruction can be viewed as an adversarial attack naturally due to the following aspects. First, we constraint our DR-Attacker to replace a single word at a specific timestep to control the magnitude of the perturbation to be small enough. Second, although the local key information, e.g., a visual object word is destroyed, the human, which is able to comprehend the long-term intention of the instruction and reasoning original instruction information according to the current visual observation, cannot be misled easily by such perturbation. However, the agent, which tends to learn the simple alignment of the instruction and visual observation, is more easily to be misled and to get stuck. Third, the replacement is conducted between words belonging to the same characteristic, ensuring the grammaticality and naturality of the original sentence. 
Since incorrect visual object, location or action words in an instruction is easy to appear in realistic scenes, e.g., a wrong annotation by human or an object previously existing but disappearing in the original scene, we impose the perturbation on visual object or location words rather than uninformative words, which can be more beneficial for enhancing the navigation robustness. 

In contrast to existing adversarial attacks on NLP which are generally static and optimized with classification-based objectives, our proposed DR-Attacker can generate dynamic perturbation on the instruction, and can be optimized by the RL paradigm under the unsupervised setting. Like other existing works which train the models on the perturbed training samples to improve the robustness of NLP models \cite{zhu2019freelb,jia2019certified,eger2019text,liu2020adversarial},  we also develop the adversarial training strategy to improve the robustness of the navigator using the perturbed instructions generated at each timestep.  Moreover, we introduce an auxiliary self-supervised reasoning task during the adversarial training stage, which can better promote the adversarial training results.



\subsection{Adversarial Attacks in Navigation}
	Although adversarial attacks are popular in verifying and improving the robustness of the deep learning models in both image \cite{salman2019provably,feng2019learning,cemgil2020adversarially,zhang2019defense} and NLP  \cite {jones2020robust,yin2020on,araujo2020on,li2020bert,zang2020word,wang2019improving,cheng2019robust,ebrahimi2018on,ren2019generating,zhu2019freelb} domains, there are few works attempting to employ the adversarial attacks for improving the robustness of the   embodied navigation agents, since the setting and environment in navigation is usually dynamic and complex. \cite{liu2020spatiotemporal} took the first attempt to introduce spatio-temporal perturbations on the visual objects for embodied question answering (EQA) task \cite{das2018embodied}, by perturbing the physical properties (e.g., texture or shape) of visual objects. They used the available ground-truth labels to guide the perturbation generation by using classification-based objectives. 
	Compared with the collection of diverse visual environments to improve the robustness of the agent, annotating large-amount of high-quality and informative instruction is more difficult and labor-intensive for the natural language grounded visual navigation task.  Therefore, in contrast to \cite{das2018embodied}, we make the first attempt to introduce adversarial attacks for the existing available instruction data in this paper, to mitigate the scarcity of available high-quality instructions which largely limits the navigation performance of existing instruction-following agents. Moreover, our introduced perturbation can be optimized in an unsupervised way, which is more practical.
	
		\subsection{
		Automatic Data Augmentation}
		Automatic data augmentation aims to learn data augmentation strategies automatically according to the target model performance instead of designing  augmentation strategies manually based on the expertise knowledge.  AutoAugment \cite{cubuk2019autoaugment}  formulates the automatic augmentation
        policy search as a discrete search problem and employs a reinforcement learning (RL) framework to search
        the policy consisting of possible augmentation operations. However,   high computational cost is required for training and evaluating thousands of sampled
        policies in the search process.
        To speed up policy search, many variants of AutoAugment are proposed \cite{ho2019population, lim2019fast,liu2020adversarial, hataya2020faster, cubuk2020randaugment}. PBA \cite{ho2019population} introduces population-based
        training to efficiently train the network parallelly across different CPUs or GPUs. Fast AutoAugment \cite{lim2019fast} moves the costly search stage from training
        to evaluation through bayesian optimization. Adversarial AutoAugment \cite{liu2020adversarial} directly learns augmentation policies on target tasks and develops an adversarial framework to jointly optimize target network training and augmentation
        policy search. The most related work to our proposed method is Adversarial AutoAugment \cite{liu2020adversarial}, where the policy sampler and the target model are jointly optimized in an adversarial way. The difference between our method and Adversarial AutoAugment is that our augmented samples are generated through the adversarial attack rather than the composition of augmentation strategies, which is constrained to be small in magnitude while impact the agent performance largely.
	
	\section{Method}
	\label{method}
	In this section, we describe the natural language grounded visual navigation task first and then introduce our proposed method.
    The problem setup is given in Sec. \ref{Problem-Setup}. The details of our proposed Dynamic Reinforced Instruction Attacker (DR-Attacker), including the optimization of the perturbation generation, the adversarial training with the auxiliary self-supervised reasoning task, and the model details are presented in Sec. \ref{language-instruction-poisoning}. 
	\subsection{Problem Setup}
	\label{Problem-Setup}
	Natural language grounded visual navigation task requires a navigator to find a route (a sequence of viewpoints) from a start viewpoint 
    to the target viewpoint 
    following the given instruction $\mathbf{I}$. For the NDH task, the  instruction $\mathbf{I}$ is composed of  
$<t_{0},\mathbf{Q}_{1},\mathbf{H}_{1},\mathbf{Q}_{2},\mathbf{H}_{2},...,\mathbf{Q}_{t},\mathbf{H}_{t}>$, which includes the given target object $t_{0}$, the questions $\mathbf{Q}$ and the answers $\mathbf{H}$ till the turn $t$ (0 $\le$ $t$ $\le$ $T$, where $T$ is the total number of question-answer turns from the intial position to the target room). 
For the VLN task, the instruction $\mathbf{I}$ is composed of $<\mathbf{G}_{1}, \mathbf{G}_{2},...,\mathbf{G}_{M}>$, where $\mathbf{G}_{m}$ ($1 \le m \le M$) denotes a single sentence and $M$ denotes the number of sentences. Since the $t_{0}$, $\mathbf{Q}_{t}$, $\mathbf{H}_{t}$, $\mathbf{G}_{m}$ can all be represented by word tokens, for both NDH and VLN tasks, we formulate the instruction $\mathbf{I}$ as a set of word tokens, $\mathbf{I}=\{w_{0},...,w_{L}\}$, where $L$ is the length of the  instruction. At timestep $t$, the navigator receives a panoramic view as the visual observation. Each panoramic view is divided into 36 image views $\{o_{t,i}\}_{i=1}^{36}$, with each of views $o_{t,i}$ containing a RGB image $b_{t,i}$ accompanied with its orientation ($\theta_{t,i}^{1}$,$\theta_{t,i}^{2}$), where $\theta_{t,i}^{1}$ and $\theta_{t,i}^{2}$ are the
angles of heading and elevation, respectively. 
We follow the \cite{tan2019learning} to obtain the view feature $\mathbf{f}_{t,i}^{v}$.
Regarding the visual observations and instructions, the navigator infers the action for each step $t$ from the candidate actions list, which consists of $J$ neighbours of the current node in the navigation graph and a stop action. Generally, the navigator is a sequence-to-sequence model with the encoder-decoder architecture \cite{anderson2018vision,thomason2019vision}.
	\subsection{Dynamic Reinforced Instruction Attacker}	
    \label{language-instruction-poisoning}
	\subsubsection{Perturbation Generation as an RL Problem}
	Since there is no direct label as that in the classification-based tasks \cite{araujo2020on,ren2019generating} for judging the success of attack in such navigation tasks, we use a reinforcement learning (RL) framework to formulate the perturbation generation. The framework contains two major components: an environment model $\mathbf{E_{\mu}}$ which is a well-trained navigator (also called  as victim navigator), and an instruction attacker $\mathbf{\pi_{\phi}}$, which can be viewed as the policy agent. The attacker $\mathbf{\pi_{\phi}}$ learns to disturb the correct action decision of $\mathbf{E_{\mu}}$ by generating perturbed instruction $\mathbf{I}_{t}'$ for $\mathbf{E_{\mu}}$ at each timestep $t$. $\mu$ and $\phi$ denote the parameters of the environment model and attacker, respectively.
	Under the RL setting, the state $s_{t} \in S$ is the visual state  $\mathbf{f}_{t}^{v}$.  The action $a_{t}\in {A}$ is the perturbation operation by substituting the selected target word in the original instruction with a candidate  word. The construction details of the target word set and candidate substitution word set for each instruction are given in Sec. \ref{construction}. Note that the attack operation is sequentially conducted at each navigation step $t$ rather than once at the beginning since the key  instruction information changes dynamically during the navigation process. 
	
	To measure the success of the attack and design reasonable reward for optimizing the attacker in such navigation tasks, we propose ``deviation from the target position'' as a metric. That is, the goal of the attacker is to enforce the navigator to make the wrong navigation trajectory and stop at a position which is far from the target position.  Therefore, the reward $r_{t}$ will be negative for the attacker if the victim navigator stops within $Z$ meters around the target viewpoint at the final step, otherwise the reward will be positive. $Z$ is a predefined distance threshold. We also adopt a direct reward \cite{wu2018building} at each non-stop step $t$ by considering the progress, i.e., the change of the distance to the target viewpoint  made by current timestep. If the navigator makes  positive progress to the target position at non-stop step $t$, the direct reward $r_{t}$ will be negative. Similar to \cite{tan2019learning}, the reward in our RL setting is set as a predefined constant.  To satisfy the `small perturbation' principle of adversarial samples \cite{ren2019generating,zang2020word,zhu2019freelb,araujo2020on}, the attacker is required to substitute only one word in the instruction at each  timestep.  

Without the loss of generality, we apply the Advantage Actor-Critic (A2C) \cite{mnih2016asynchronous} algorithm to iteratively update the parameters of the attacker $\mathbf{\pi_{\phi}}$. A2C framework contains a policy network $\pi(\mathbf{s}|\phi_{\pi})$ (here is the attacker) and a value network  $V(\mathbf{s}|\phi_{v})$ to learn a optimal policy. $\phi_{\pi}$ and $\phi_{v}$ denote the parameters of the network. Given the {\it state-action-reward} $(s_{t}, a_{t}, r_{t})$ of $\forall t \in (0, N)$ observation at each step $t$, the algorithm computes the total accumulated reward $R_{t}$, the policy gradient $\nabla_{pg}$, the value gradient $\nabla_{v}$ and the entropy gradient $\nabla_{h}$ by: 
\begin{align}
R_{t}&=\sum\nolimits_{i=t}^{N}\gamma^{i-t}r_{t}+\gamma^{N-t}V(s_{N+1}),\\
\nabla_{pg}&=\nabla_{\phi_{\pi}}\mathrm{log}(\pi(s_{t},a_{t}|\phi_{\pi}))A_{t},\\
\nabla_{v}&=\frac{\partial(V(s_{t}|\phi_{v})-R_{t})^2}{\partial\phi_{v}},\\
\nabla_{h}&=\nabla_{\phi_{\pi}}\sum\nolimits_{i=0}^{N}\mathrm{log}(\pi(s_{i},a_{i}|\phi_{\pi}))\pi(s_{i},a_{i}|\phi_{\pi}).
\end{align}
where $\gamma \in [0,1)$ is the discount factor. $A_{t}=R_{t}-V(s_t)$ is the advantage. 
Subsequently, an optimization step is performed in the direction that maximizes both $\mathbb{E}[R_{t}]$ (direction $\nabla_{pg}$) and the entropy of $\pi(s_{t})$ (direction $\nabla_{h}$), as well as minimizes the mean squared error of $V(s_{t})$ (direction -$\nabla_{v}$).
Therefore, by using the RL paradigm, the attacker can learn to generate the perturbed  instructions at each timestep for disturbing the action decision of the navigator and misleading it to stop at the wrong target position. In our settings, the value network is a two-layer MLP.
	
	\begin{figure*}
	\centering
	\includegraphics[width=1.0\linewidth]{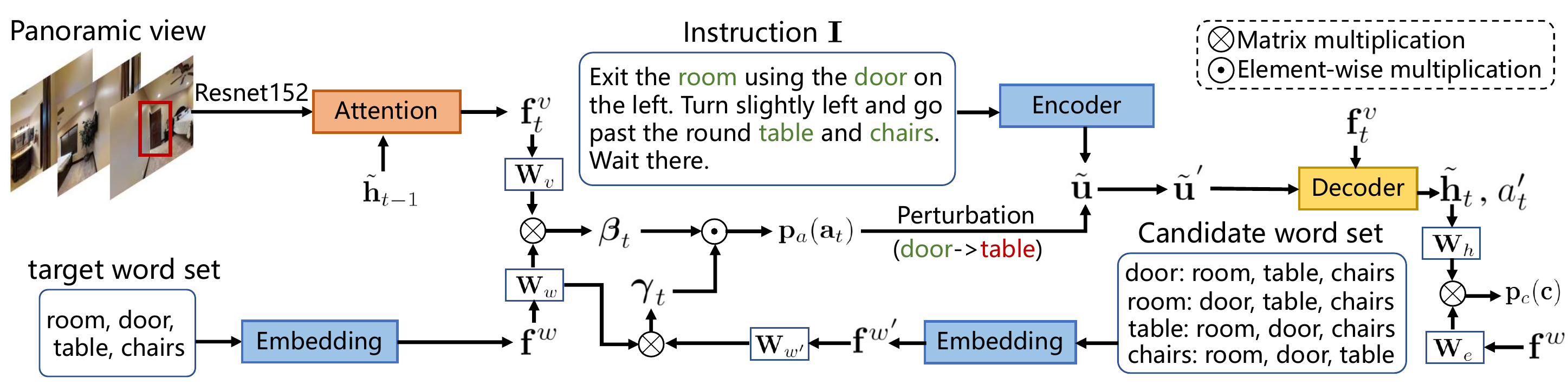}
	\caption{
		The forward processes of the DR-Attacker and the navigator. The attack score $\mathbf{p}_{a}(\mathbf{a}_t)$ is calculated by the element-wise multiplication of the word importance vector $\boldsymbol{\beta}_{t}$ and the substitution impact matrix $\boldsymbol{\gamma}_{t}$. After performing the perturbation operation on the original instruction $\tilde{\mathbf{u}}$ to generate perturbed instruction $\tilde{\mathbf{u}}^{'}$, the decoder of the navigator gets the perturbed instruction $\tilde{\mathbf{u}}^{'}$ and the attended visual feature $\mathbf{f}_{t}^{v}$ to predict the navigation action $a'_{t}$ at timestep $t$. The updated hidden state $\tilde{\mathbf{h}}_{t}$ of the decoder and the target word feature $\mathbf{f}^{w}$ are used to calculate the actual attacked word prediction probability $\mathbf{p}_{c}(\mathbf{c})$.  
	}
	\label{fig:forward process}
	\vspace{-0.2cm}
\end{figure*}
	
\subsubsection{Adversarial Training with Auxiliary Self-supervised Task}	
For improving the navigation robustness, we develop an effective adversarial training strategy, which can encourage the joint optimization for the victim navigator and the attacker. 
Through alternative optimization under the adversarial setting, the attacker can iteratively learn to create misleading instructions for confusing the victim navigator, while the victim navigator is trained on the perturbed instructions to enhance the model robustness.
Motivated by \cite{pinto2017robust}, we use the RL strategy for training both the victim navigator and the attacker, and formulate the adversarial setting as the two-player zero-sum Markov games. At each timestep $t$, both the attacker and the victim navigator receive the visual observation $\mathbf{f}_{t}^{v}$ and the language instruction $\mathbf{I}_{t}$ ($\mathbf{I}_{t}'$ for the navigator, $\mathbf{I}_{t}$ is invariant while $\mathbf{I}_{t}'$ is variable). Then the attacker takes the action by generating the perturbed instruction, and the navigator takes the action by moving to the next viewpoint.  With the inverse objective of the navigation, i.e., the navigator is supposed to stop at the nearest point from the target position, an inverse reward is set for the  attacker and the navigator: $r_{\pi}=-r_{\eta}$ ($r_{\eta}$ is represented by $R_{Nav}$ in Figure \ref{fig:framework}), where $\pi$ and $\eta$ represent the policies for the attacker and the victim navigator, respectively. Therefore, our adversarial setting can be represented by:
\begin{equation}
\begin{aligned}
r_{\eta}^{*}=\mathop{\mathrm{min}}\limits_{\pi}\mathop{\mathrm{max}}\limits_{\eta}r_{\eta}(\eta,\pi).
\end{aligned}
\end{equation}   
We conduct the alternative optimization procedure between the navigator and the attacker, namely, keep the parameters of one agent fixed and optimize another. The optimization procedure of adversarial training is given in Algorithm \ref{algorithm}. At stage 1, we pre-train the navigator and use the pre-trained navigator to pre-train the attacker. At stage 2, we conduct alternative iteration procedure between the navigator and the attacker to implement the joint optimization. For facilitating implementation, the RL strategy for training the victim navigator also follows the A2C algorithm which was similar to  \cite{tan2019learning}. 

To encourage the agent to capture actual key information and improve the fault-tolerance ability with perturbed instructions, which is important for robust navigation, we introduce an auxiliary self-supervised reasoning task during the training phases of the victim navigator, by asking the navigator to predict the actual attacked word by the attacker at each timestep $t$:
\begin{equation}
\begin{aligned}
\mathbf{p}_{c}(\mathbf{c}) = \mathrm{Softmax}((\mathbf{f}^{w}\mathbf{W}_{e})(\tilde{\mathbf{h}}_{t}\mathbf{W}_{h})^{T}).
\end{aligned}
\end{equation} 
where $\mathbf{c}$ is the target word set for the given instruction $\mathbf{I}$ and  $\mathbf{p}_{c}(\mathbf{c})$ denotes the prediction probability. $\mathbf{f}^{w} \in \mathbb{R}^{L'\times D_{w}}$ denotes the target word features. $L'$ is the size of the target word set.  $\tilde{\mathbf{h}}_{t} \in \mathbb{R}^{1\times D_h}$ represents the visual-and-instruction aware hidden state feature of the decoder \cite{tan2019learning}  in the navigator. $\mathbf{W}_{e} \in \mathbb{R}^{D_{w}\times D_{p}}$ and $\mathbf{W}_{h} \in \mathbb{R}^{D_{h}\times D_{p}}$ denote the learnable linear transformations. $D_{w}$, $D_{h}$ and $D_{p}$ denote the feature dimensions. 
The prediction is optimized by cross-entropy loss and the ground-truth label is the actual attacked word by the attacker. 
As a result, the probability that 
the agent captures the actual important instruction information and haves the self-correction ability can be increased with the accuracy improvement of the attacker for attacking key instruction information.  Therefore, through the auxiliary self-supervision reasoning task, the enhancement of the attacker can effectively lead to the improvement of the navigator.

\begin{algorithm}[t]
	\SetAlgoLined
	\caption{Adversarial Training}
	\label{algorithm}
	{\small{
			\KwIn{the navigator NAV with policy $\eta$, the attacker ATT with policy $\pi$}
			\KwOut{Optimized parameters $\theta_{N_{iter}}^{\eta}$ for $\eta$ and $\theta_{N_{iter}}^{\pi}$ for $\pi$}	
			\BlankLine
			// Stage 1: Initialization 
			\BlankLine
			Pre-train NAV with original training set to get $\theta_{0}^{\eta}$ \\
			Pre-train ATT with the pretrained NAV of fixed parameters $\theta_{0}^{\eta}$ to get
			$\theta_{0}^{\pi}$
			\BlankLine
			// Stage 2: Adversarial training 
			\BlankLine
			\For{$i=1:N_{iter}$}
			{
				// Fix $\theta^{\pi}$ to optimize $\theta^{\eta}$\\
				$\theta_{i}^{\eta}$ $\leftarrow$ $\theta_{i-1}^{\eta}$\\
				\For{$j=1:N_{\eta}$}
				{
					\{($s_{t}$,$a_{t}^{\eta}$,$r_{t}^{\eta}$)\} $\leftarrow$ rollout($\mathbf{I}_{t}'$, $f_{t}^{v}$, $\eta_{\theta_{i}^{\eta}}$, $\pi_{\theta_{i-1}^{\pi}}$)\\
					$\theta_{i}^{\eta}$ $\leftarrow$ policyOptmizer(\{($s_{t}$,$a_{t}^{\eta}$, $r_{t}^{\eta}$)\}, $\eta$, $\theta_{i}^{\eta}$)\\
				}
				// Fix $\theta^{\eta}$ to optimize $\theta^{\pi}$\\
				$\theta_{i}^{\pi}$ $\leftarrow$ $\theta_{i-1}^{\pi}$\\
				\For{$j=1:N_{\pi}$}
				{
					\{($s_{t}$,$a_{t}^{\pi}$,$r_{t}^{\pi}$)\} $\leftarrow$ rollout($\mathbf{I}_{t}$, $f_{t}^{v}$, $\eta_{\theta_{i}^{\eta}}$, $\pi_{\theta_{i}^{\pi}}$)\\
					$\theta_{i}^{\pi}$ $\leftarrow$ policyOptmizer(\{($s_{t}$, $a_{t}^{\pi}$, $r_{t}^{\pi}$)\}, $\pi$, $\theta_{i}^{\pi}$)\\
				}
			}
			\Return $\theta_{N_{iter}}^{\eta}$, $\theta_{N_{iter}}^{\pi}$
	}}
	
\end{algorithm}

	\subsubsection{Model Details}
	\label{construction}

\begin{figure*}
	\centering
	\includegraphics[width=1.0\linewidth]{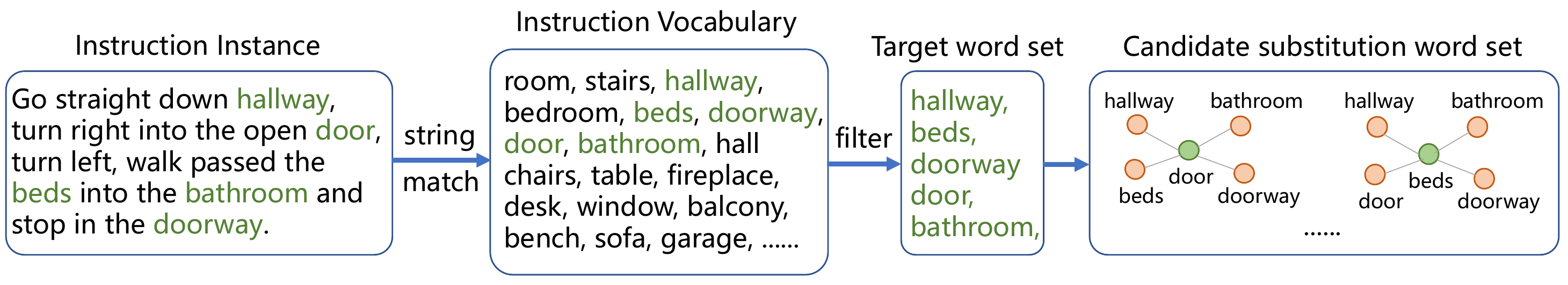}
	\caption{
		The construction details of target word set and candidate substitution word set on VLN. The target word set is constructed for each instruction by conducting string match between the instruction and the instruction vocabulary which only contains words indicating visual objects and locations. The candidate substitution word set for each target word is built by  collecting the remaining target words in the same instruction. 
	}
	\vspace{-0.2cm}
	\label{fig:VLN-construction-details}
\end{figure*}
\begin{figure*}
	\centering
	\includegraphics[width=1.0\linewidth]{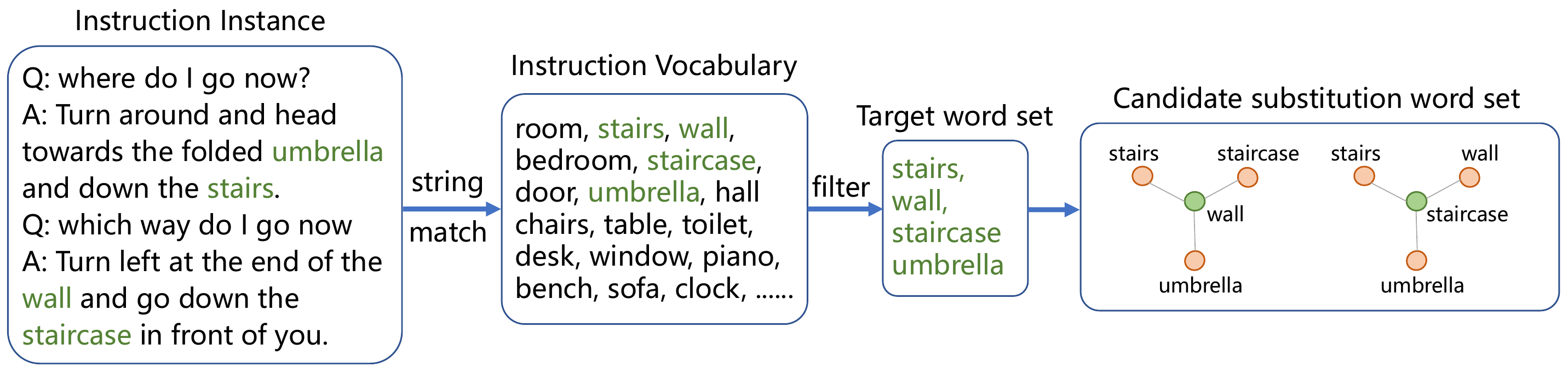}
	\caption{
		The construction details of target word set and candidate substitution word set on NDH. Since the last answer in the instruction generally contains the guiding information, we only construct the target word set and perform the perturbation operation for the last answer in the instruction for each instance.
	}
	\label{fig:NDH-construction-details}
	\vspace{-0.4cm}
\end{figure*}
	
	\textbf{Forward Process of the Instruction Attacker.}
    In this part, we describe the forward process of the proposed DR-Attacker, i.e., the attacker $\mathbf{\pi_{\phi}}$ in detail.  At each timestep $t$, the DR-Attacker calculates the action prediction probability, also referred to as the attack score, by considering both the word importance in the current instruction and the substitution impact of different candidate words (illustrated in Figure \ref{fig:framework}). 
Within the prior that the words indicate visual object (e.g., ``door'') and location (e.g., ``bathroom'') are   most informative for guiding the navigation, we construct the target word set by selecting these two kinds of words for each instruction in advance.
For target word $w_{j}$ ($0\le j\le L'$, $L'$ is the size of target word set) in the instruction $\mathbf{I}$, we denote the candidate substitution word set of $w_{j}$ as $\{w'_{j,k}\}_{k=1}^{K}$, where $K$ is the size of candidate substitution word  set. 
To promote the understanding of the given instruction as well as keep a reasonable set size,  we choose the remained target words in the same instruction to construct the candidate substitution word set for the specific target word. At timestep $t$, a word importance vector $\boldsymbol{\beta}_{t} \in \mathbb{R}^{L'}$ is first caculated by:
\begin{equation}
\begin{aligned}
\boldsymbol{\beta}_{t}=\mathrm{Softmax}((\mathbf{f}^{w}\mathbf{W}_{w})(\mathbf{f}_{t}^{v}\mathbf{W}_{v})^{T}),
\end{aligned}
\end{equation}   
where $\mathbf{f}^{w} \in \mathbb{R}^{L'\times D_{w}}$ and $\mathbf{f}_{t}^{v} \in \mathbb{R}^{1\times D_{v}}$ represent the word features encoded by BiLSTM of target words and attended visual feature \cite{tan2019learning}, respectively. $\mathbf{W}_{w} \in \mathbb{R}^{D_{w}\times D_{p}}$ and $\mathbf{W}_{v} \in \mathbb{R}^{D_{v}\times D_{p}}$ are the learnable linear transformations that convert the different features into the same embedding space. $D_{w}$, $D_{v}$ and $D_{p}$ represent the feature dimensions. Then, the substitution impact of different candidate words for each target word $w_{j}$ is obtained by:
\vspace{-0.2cm}
\begin{equation}
\begin{aligned}
\boldsymbol{\gamma}_{t,j}=\mathrm{Softmax}((\mathbf{f}_{j}^{w}\mathbf{W}_{w})(\mathbf{f}_{j}^{w'}\mathbf{W}_{w'})^{T}),
\end{aligned}
\end{equation}
where $\mathbf{f}_{j}^{w} \in \mathbb{R}^{1\times D_{w}}$ and $\mathbf{f}_{j}^{w'} \in \mathbb{R}^{K\times D_{w}}$ denote the word features of target word $w_{j}$ and candidate words $w'_{j}$. $\mathbf{W}_{w'} \in \mathbb{R}^{D_{w}\times D_{p}}$ is the learnable linear transformation. After calculating the substitution impact of different candidate words for all the target words in the instruction to obtain the substitution impact matrix $\boldsymbol{\gamma}_{t} \in \mathbb{R}^{L'\times K}$, the attack score $\mathbf{p}_{a}(\mathbf{a}_t) \in \mathbb{R}^{L'\times K}$, i.e., the action prediction probability of the DR-Attacker is calculated by: 
\vspace{-0.1cm}
\begin{equation}
\begin{aligned}
\mathbf{p}_{a}(\mathbf{a}_t)=\mathrm{Softmax}(\boldsymbol{\beta}_{t}\circ \boldsymbol{\gamma}_{t}).
\end{aligned}
\end{equation}
where $\circ$ denotes the element-wise multiplication. $\mathbf{a}_{t}$ represents the  candidate action set with the size of $L'\times K$.  Through the learnable attack score $\mathbf{p}_{a}(\mathbf{a}_t)$, the DR-Attacker can learn to generate the optimal perturbation at each timestep $t$. Note that while there will be a semantic change compared with the original target word based on our word substitution strategy, we do not distinguish the perturbed instruction with the conventional adversarial samples. This is because the impact of single word substitution is subtle on the overall intention of whole instruction. 
	
\renewcommand{\arraystretch}{1.2}
\begin{table*}[!htb]
	\fontsize{9}{9}\selectfont
	\caption{The comparison results with state-of-the-art methods on R2R dataset. Apart from NE, higher value indicates better results.}
	\vspace{-0.2cm}
	\label{tab:comparison results on VLN}
	\begin{center}
		\begin{tabular}{|c|c|c|c|c|c|c|c|c|c|}
			\hline
			\multirow{2}{*}{Method}&\multicolumn{3}{c|}{Val Seen }&\multicolumn{3}{c|}{Val Unseen}&\multicolumn{3}{c|}{Test Unseen}\cr\cline{2-10}
			&NE(m) $\downarrow$&SR(\%) $\uparrow$&SPL(\%)  $\uparrow$&NE(m) $\downarrow$&SR(\%)  $\uparrow$&SPL(\%) $\uparrow$&NE(m) $\downarrow$&SR (\%) $\uparrow$&SPL(\%) $\uparrow$\cr
			\hline
			seq-2-seq \cite{anderson2018vision} &6.01&39&-&7.81&22&-&7.85&20&18 \\
			Speaker-Follower \cite{fried2018speaker}&3.36&66&-&6.62&35&-&6.62&35&28 \\
			
			Regretful \cite{ma2019the}&\textbf{3.23}&69&63&5.32&50&41&5.69&48&40 \\
			RCM \cite{wang2019reinforced}&3.53&67&-&6.09&43&-&6.12&43&38 \\
			PRESS \cite{li2019robust}&4.39&58&55&5.28&49&45&5.59&49&45 \\
			EnvDrop \cite{tan2019learning}&3.99&62&59&5.22&52&48&\textbf{5.23}&51&47 \\
			Ours&3.52&\textbf{70}&\textbf{67}&\textbf{4.99}&\textbf{53}&\textbf{48}&5.53&\textbf{52}&\textbf{49} \\
			
			\hline
		\end{tabular}
	\end{center}
\end{table*}

\renewcommand{\arraystretch}{1.5}
\begin{table*}[!htb]
	\fontsize{9}{9}\selectfont
	\caption{The comparison results with state-of-the-art methods on CVDN dataset. The Goal Progress (GP) (m) is reported following most existing works.  
	}
	\vspace{-0.2cm}
	\label{tab:comparison results on NDH}
	\begin{center}
		\begin{tabular}{|c|c|c|c|c|c|c|c|c|c|}
			\hline
			\multirow{2}{*}{Method}&\multicolumn{3}{c|}{Val Seen }&\multicolumn{3}{c|}{Val Unseen}&\multicolumn{3}{c|}{Test Unseen}\cr\cline{2-10}
			&Oracle&Navigator&Mixed&Oracle&Navigator&Mixed&Oracle&Navigator&Mixed\cr
			\hline

			sequence-to-sequence \cite{thomason2019vision}&4.48&5.67&5.92&1.23&1.98&2.10&1.25&2.11&2.35 \\
			CMN \cite{zhu2020vision}&5.47&6.14&7.05&2.68&2.28&2.97&2.69&2.26&2.95 \\ 
		PREVALENT \cite{hao2020towards}&-&-&-&2.58&2.99&3.15&1.67&2.39&2.44 \\	Ours&\textbf{5.60}&\textbf{7.58}&\textbf{8.06}&\textbf{3.27}&\textbf{4.00}&\textbf{4.18}&\textbf{2.77}&\textbf{2.95}&\textbf{3.26} \\
			
			\hline
			
		\end{tabular}
	\vspace{-0.2cm}
	\end{center}
\end{table*}

\renewcommand{\arraystretch}{1.5}
\begin{table*}[!htb]
	\fontsize{9}{9}\selectfont
	\caption{The comparison of training time, data and device between PREVALENT \cite{hao2020towards} and our method on NDH.
	}
	\vspace{-0.2cm}
	\label{tab:comparison-PREVALENT-ours}
	\begin{center}
		\begin{tabular}{|c|c|c|c|c|c|c|c|}
			\hline
			\multirow{2}{*}{Method}&\multicolumn{3}{c|}{Time (min) }&\multicolumn{2}{c|}{Data}&\multicolumn{2}{c|}{Device}\cr\cline{2-8}
			&Pretrain&Other phases&Total&Pretrain&Other phases&Pretrain&Other phases\cr
			\hline

		    PREVALENT \cite{hao2020towards}&-&1661&-&6, 582, 000&4, 742&8 v100 GPUs& 1 1080Ti GPU\\
		    Ours&143&328&471&4, 742&4, 742&1 1080Ti GPU&1 1080Ti GPU\\
			\hline
			
		\end{tabular}
	\vspace{-0.4cm}
	\end{center}
\end{table*}

	\textbf{Forward Process of the Navigator.}
    After introducing the forward process of the instruction attacker $\mathbf{\pi_{\phi}}$, we present the forward process of the navigator in this subsection.  Specifically,  the navigator follows an encoder-decoder architecture, where both the encoder and decoder are LSTMs \cite{tan2019learning}. The encoder contains a word embedding layer and a bi-directional LSTM, and its output  is the language feature $\{\tilde{\mathbf{u}}_{l}\}_{l=1}^{L}$ of the instruction:
\begin{equation}
\begin{aligned}
\mathbf{u}_{l}=&\mathrm{Embedding}(w_{l}), \\
\tilde{\mathbf{u}}_{1}, \tilde{\mathbf{u}}_{2},..., \tilde{\mathbf{u}}_{L}=& \mathrm{Bi}\mathrm{-}\mathrm{LSTM}(\mathbf{u}_{1},\mathbf{u}_{2},...\mathbf{u}_{L}).
\end{aligned}
\end{equation} 
Then, the decoder receives the attended visual feature $\mathbf{f}_{t}^{v}$ and language feature $\tilde{\mathbf{u}}$, and generates the visual-and-instruction aware hidden state $\tilde{\mathbf{h}}_{t}$:
\begin{align}
\mathbf{h}_{t}=&\mathrm{LSTM}([\mathbf{f}_{t}^{v};\mathbf{a}'_{t-1}],\tilde{\mathbf{h}}_{t-1}), \\
\alpha_{t,l}^{w}=&\mathrm{Softmax}(\tilde{\mathbf{u}}_{l}\mathbf{W}_{u}\mathbf{h}_{t}), \\
\mathbf{f}_{t}^{w}=&\sum_{l}\alpha_{t,l}^{w}\tilde{\mathbf{u}}_{l}, \\
\tilde{\mathbf{h}}_{t}=&\mathrm{tanh}(\mathbf{W}_{h'}[\mathbf{f}_{t}^{w};\mathbf{h}_{t}]),
\end{align}
where $\mathbf{a}'_{t-1}$ is the action feature of the timestep $t-1$.  $\mathbf{W}_{u} \in R^{D_{w}\times D_{h}}$ and $\mathbf{W}_{h'} \in R^{(D_{w}+D_{h})\times D_{h}}$ are the learnable linear transformations. The attended visual feature $\mathbf{f}_{t}^{v}$ is calculated by:
\begin{align}
\alpha_{t,i}^{v}=&\mathrm{Softmax}(\mathbf{f}_{t,i}^{v}\mathbf{W}_{v'}\tilde{\mathbf{h}}_{t-1}),\\
\mathbf{f}_{t}^{v}=&\sum_{i}\alpha_{t,i}^{v}\mathbf{f}_{t,i}^{v},
\end{align}
where $\mathbf{W}_{v'} \in R^{D_{v}\times D_{h}}$ is the learnable linear transformation.
Then, the action prediction probability $\mathbf{p}_{n}(\mathbf{a}'_{t})$ of the navigator is calculated by:
\begin{equation}
\begin{aligned}
\mathbf{p}_{n}(\mathbf{a}'_{t})=&\mathrm{Softmax}(\mathbf{c}'_{t,k}\mathbf{W}_{a}\tilde{\mathbf{h}}_{t}),
\end{aligned}
\end{equation}
where $\mathbf{c}'_{t,k}$ denotes the candidate action features. $\mathbf{W}_{a} \in R^{D_{v}\times D_{h}}$ is the trainable linear transformation. The navigator takes the action $a'_{t}$ according to the action prediction probability $\mathbf{p}_{n}(\mathbf{a}'_{t})$. 

The forward processes of the attacker and the navigator are shown in Figure \ref{fig:forward process}. As illustrated in Figure \ref{fig:forward process}, based on the attack score $\mathbf{p}_{a}(\mathbf{a}_t)$ which is calculated by the elementwise multiplication of word importance vector $\boldsymbol{\beta}_{t}$ and substitution impact matrix $\boldsymbol{\gamma}_{t}$, the perturbation operation is conducted on the original instruction $\tilde{\mathbf{u}}$ to generate perturbed instruction $\tilde{\mathbf{u}}^{'}$. Then the decoder receives the attended visual feature $\mathbf{f}_{t}^{v}$ and the perturbed instruction $\tilde{\mathbf{u}}^{'}$ to predict the next action $a'_{t}$. The updated hidden state $\tilde{\mathbf{h}}_{t}$ of the decoder and the target word feature $\mathbf{f}^{w}$ are used to calculate the prediction probability $\mathbf{p}_{c}(\mathbf{c})$ of the actual attacked word  for the self-supervised auxiliary reasoning task.

	\textbf{Construction Details of Target Word Set and Candidate Word Set.}
	In this part, we show the construction details of target word set and candidate substitution word set for both VLN and NDH tasks.  Specifically, for each instruction, we first construct its target word set by conducting string match between it and the instruction vocabulary. The instruction vocabulary contains the words indicating visual objects or locations, which are collected from the given instruction vocabulary from the dataset. Then, the candidate substitution word set is constructed for each target word by selecting the remained target words in the same instruction. The construction details of the  target word set and candidate substitution word set for VLN and NDH tasks are shown in Figure  \ref{fig:VLN-construction-details} and Figure \ref{fig:NDH-construction-details}, respectively. Note that since the last answer in the dialog history  plays the direct role of guiding navigation in the NDH task, we only construct the target word set and conduct the perturbation for the last answer in the NDH task, as shown in Figure \ref{fig:NDH-construction-details}.
	
\renewcommand{\arraystretch}{1.2}
\begin{table*}[!htb]
	\fontsize{9}{9}\selectfont
	\caption{The ablation study results on R2R dataset. NE (m), SR (\%), SPL (\%) results are reported. Apart from NE, higher value indicates better results. }\label{tab:ablation1}
	\vspace{-0.2cm}
	\begin{center}
		\begin{tabular}{|c|c|c|c|c|c|c|}
			\hline
			\multirow{2}{*}{Method}&\multicolumn{3}{c|}{Val Seen }&\multicolumn{3}{c|}{Val Unseen}\cr\cline{2-7}
		&NE (m) $\downarrow$&SR (\%) $\uparrow$&SPL (\%) $\uparrow$&NE (m) $\downarrow$&SR (\%) $\uparrow$&SPL (\%) $\uparrow$\cr
			\hline
			Base Agent &4.37&58.7&56&5.43&48.0&45 \\
			DR-Attacker&4.55&57.3&55&5.62&46.6&43 \\
			Adversarial Training w auxiliary task&4.15&62.0&59&5.25&49.6&46
			\\
			Finetune&3.52&70.2&67&4.99&53.2&48 \\
			
			\hline
		\end{tabular}
	\end{center}
\end{table*}

\renewcommand{\arraystretch}{1.2}
\begin{table*}[!htb]
	\fontsize{9}{9}\selectfont
	\caption{The ablation study results on CVDN dataset. The Goal Progress (GP) (m) is reported. The supervision setting is mixed supervision.}
	\vspace{-0.2cm}
	\label{tab:ablation2}
	\begin{center}
		\begin{tabular}{|c|c|c|c|c|c|c|}
			\hline
			\multirow{2}{*}{Settings}&\multicolumn{3}{c|}{Val Seen }&\multicolumn{3}{c|}{Val Unseen }\cr\cline{2-7}
			&Last A&Last QA&All&Last A&Last QA&All\cr
			\hline
			
			Base Agent &7.55&7.15&7.35&3.88&3.95&3.72\\
			DR-Attacker&5.30&5.88&5.85&1.95&2.51&2.29\\
			Adversarial Training w auxiliary task&6.80&6.96&7.23&3.90&3.80&3.93\\
			Finetune&7.66&7.61&8.06&4.20&4.19&4.18\\
			\hline


		\end{tabular}
	\end{center}
\end{table*}

\renewcommand{\arraystretch}{1.2}
\begin{table*}[!htb]
	\fontsize{9}{9}\selectfont
	\caption{The ablation study results on CVDN dataset. The Goal Progress (GP) (m) is reported. The dialog history setting is last answer.}
	\vspace{-0.2cm}
	\label{tab:DR-Attacker and Adversarial Training on NDH}
	\begin{center}
		\begin{tabular}{|c|c|c|c|c|c|c|}
			\hline
			\multirow{2}{*}{Settings}&\multicolumn{3}{c|}{Val Seen }&\multicolumn{3}{c|}{Val Unseen }\cr\cline{2-7}
			&Oracle&Navigator&Mixed&Oracle&Navigator&Mixed\cr
			\hline
			
			Base Agent &5.44&6.92&7.55&3.28&4.06&3.88\\
		DR-Attacker&4.00&5.07&5.30&1.50&2.38&1.95\\	Finetune (Adversarial Training) &5.48&7.13&7.61&3.37&4.16&4.08\\
			Finetune (Adversarial Training w auxiliary task)&5.52&7.49&7.66&3.48&4.21&4.20\\
			\hline


		\end{tabular}
	\vspace{-0.2cm}
	\end{center}
\end{table*}

\renewcommand{\arraystretch}{1.5}
\begin{table}[!htb]
	\fontsize{9}{9}\selectfont
	\caption{The comparison results of different adversarial attacking mechanisms in attacking and promoting the navigation performance on CVDN dataset. The Goal Progress (GP) (m) is reported. The dialog history setting is last answer. Ora., Nav., and Mix. represent the supervision is Oracle, Navigator and Mixed, respectively.
	}
	\vspace{-0.2cm}
	\label{tab:adv on NDH}
	\begin{center}
		\begin{tabular}{|c|c|c|c|c|c|c|}
			\hline
			\multirow{2}{*}{Method}&\multicolumn{3}{c|}{Val Seen }&\multicolumn{3}{c|}{Val Unseen}\cr\cline{2-7}
			&Ora.&Nav.&Mix.&Ora.&Nav.&Mix.\cr
			\hline

			 \multicolumn{7}{|c|}{Direct Attack}\\\hline
			 Static&4.09&5.32&5.48&1.71&2.63&2.23\\
			 Random&4.20&5.71&5.66&1.64&2.54&2.15\\
			 Heuristics&4.07&\textbf{4.99}&5.35&1.52&2.46&\textbf{1.89}\\
			 PWWS \cite{ren2019generating}&4.04&5.51&5.44&1.64&2.59&2.10\\
			 DR-Attacker&\textbf{4.00}&5.07&\textbf{5.30}&\textbf{1.50}&\textbf{2.38}&1.95\\\hline
		    \multicolumn{7}{|c|}{Adversarial Training}\\\hline
		    Static&5.42&6.57&7.09&3.25&3.93&3.97\\
		    Random&4.89&6.52&7.02&3.15&3.93&3.88\\
		    Heuristics&\textbf{5.60}&6.91&6.99&3.04&3.81&3.67\\
		    PWWS \cite{ren2019generating}&5.23&6.57&6.87&3.20&3.59&3.09\\
		    DR-Attacker&5.52&\textbf{7.49}&\textbf{7.66}&\textbf{3.48}&\textbf{4.21}&\textbf{4.20}\\
			\hline

			
		\end{tabular}
	\vspace{-0.4cm}
	\end{center}
\end{table}

	\section{Experiment}
	\label{experiment}
	In this section, we first introduce the datasets we use on NDH and VLN tasks, evaluation metrics, and implementation details in Sec. \ref{Experimental Setup}. Then we provide the quantitative and qualitative results in Sec. \ref{Quantitative Results} and Sec. \ref{Qualitative Results}, respectively.
	\subsection{Experimental Setup}	
	\label{Experimental Setup}
	\subsubsection{Datasets}
	\textbf{CVDN} dataset \cite{thomason2019vision} contains 2050 human-human navigation dialogs and over 7k trajectories in 83 MatterPort houses. Each trajectory is punctuated by several question-answer exchanges. Each dialog begins with an ambiguous instruction, and the subsequent dialog interaction between the navigator and oracle leads the navigator to find the target position. 
	\textbf{R2R} dataset \cite{anderson2018vision} includes 10,800 panoramic views and 7,189 trajectories. Each panoramic view has 36 images and each trajectory is paired with three natural language instructions. Both CVDN and R2R datasets are split into a training set, a seen validation set, an unseen validation set,  and a test set.

	\subsubsection{Evaluation Metrics}	The following four metrics \cite{anderson2018vision} are used for evaluation on R2R dataset: 1) Trajectory Length (TL) measures the average length of navigation trajectories in meters, 2) Navigation Error (NE)  is the distance between target viewpoint  and agent stopping position, 3) Success Rate (SR) calculates the success rate of reaching the goal, 4) Success rate weighted by Path Length (SPL) makes the trade-off between SR and TL. Based on the metrics on R2R dataset, there are some new metrics used for evaluation on CVDN dataset \cite{zhu2020vision}: 1) Goal Progress (GP) measures the average agent progress towards the goal location, 2) Oracle Success Rate (OSR) is the success rate if the agent can stop at the nearest point to the goal along its trajectory, 3) Oracle Path Success Rate (OPSR) means the success rate if the agent can stop at the  closest point to the goal along the shortest path.

	\subsubsection{Implementation Details} The navigator architecture,  training hyperparameters and the training strategy  we use in both VLN and NDH tasks are the same to \cite{tan2019learning}.   $D_{w}$, $D_{w'}$, $D_{v}$, $D_{p}$ and $D_{h}$ are set as 512, 512, 2052, 512  and 512, respectively. 
	The positive/negative rewards of the final step and each non-stop step are set as 3/-3 and 1/-1, respectively. For both VLN and NDH, we split the training  process for four steps: 1) pre-train the navigator using the original training set 2) pre-train the DR-Attacker on the pre-trained navigator and keep the parameters of the navigator fixed 3) adversarially train  both the navigator and DR-Attacker by alternative iteration 4) finetune the navigator on the original training set.  The training iterations of four steps for VLN are 40K, 10K, 40K, 200K and the training iterations of four steps for NDH are 5K, 1K, 3K, 3K. For the adversarial training, the alternation is conducted after 3K and 1K iterations for VLN and NDH, respectively. Following \cite{fried2018speaker}, we also use the data augmentation of instruction to improve the navigation performance. For improving the learning  efficiency, we also introduce imitation learning supervision \cite{tan2019learning} when training the navigator in the adversarial training stage.

   \begin{figure*}[t]
	\setlength{\abovecaptionskip}{3pt}
	\setlength{\belowcaptionskip}{3pt}
	\centering
	\renewcommand{\figurename}{Figure}
	\vspace{-0.01in}
	\subfloat[NE(m) for different scenes]{
		\includegraphics[width=5.5cm,height=4cm]{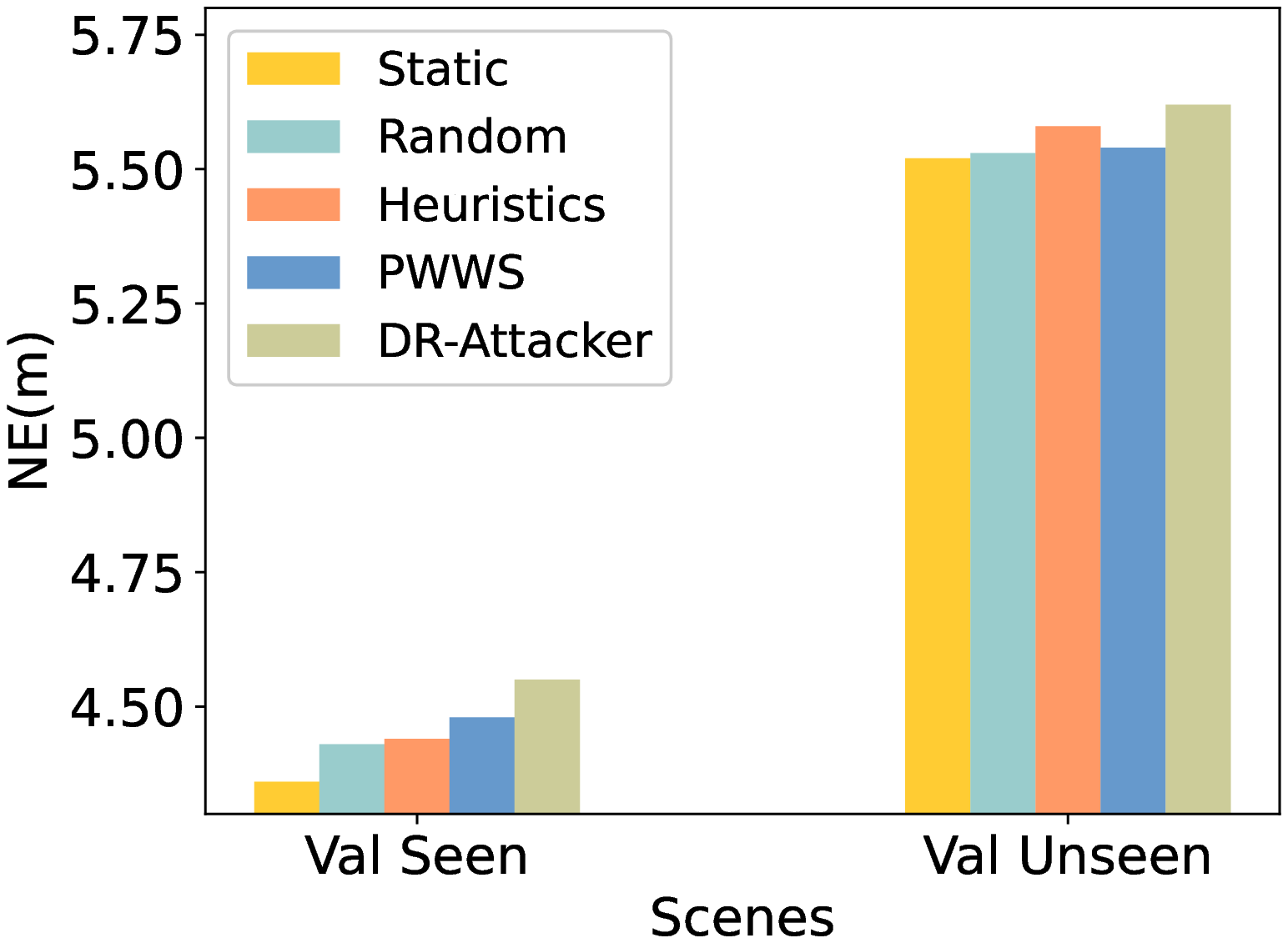}}
	\hspace{-0.01in}
	\subfloat[SR(\%) for different scenes]{
		\includegraphics[width=5.5cm,height=4cm]{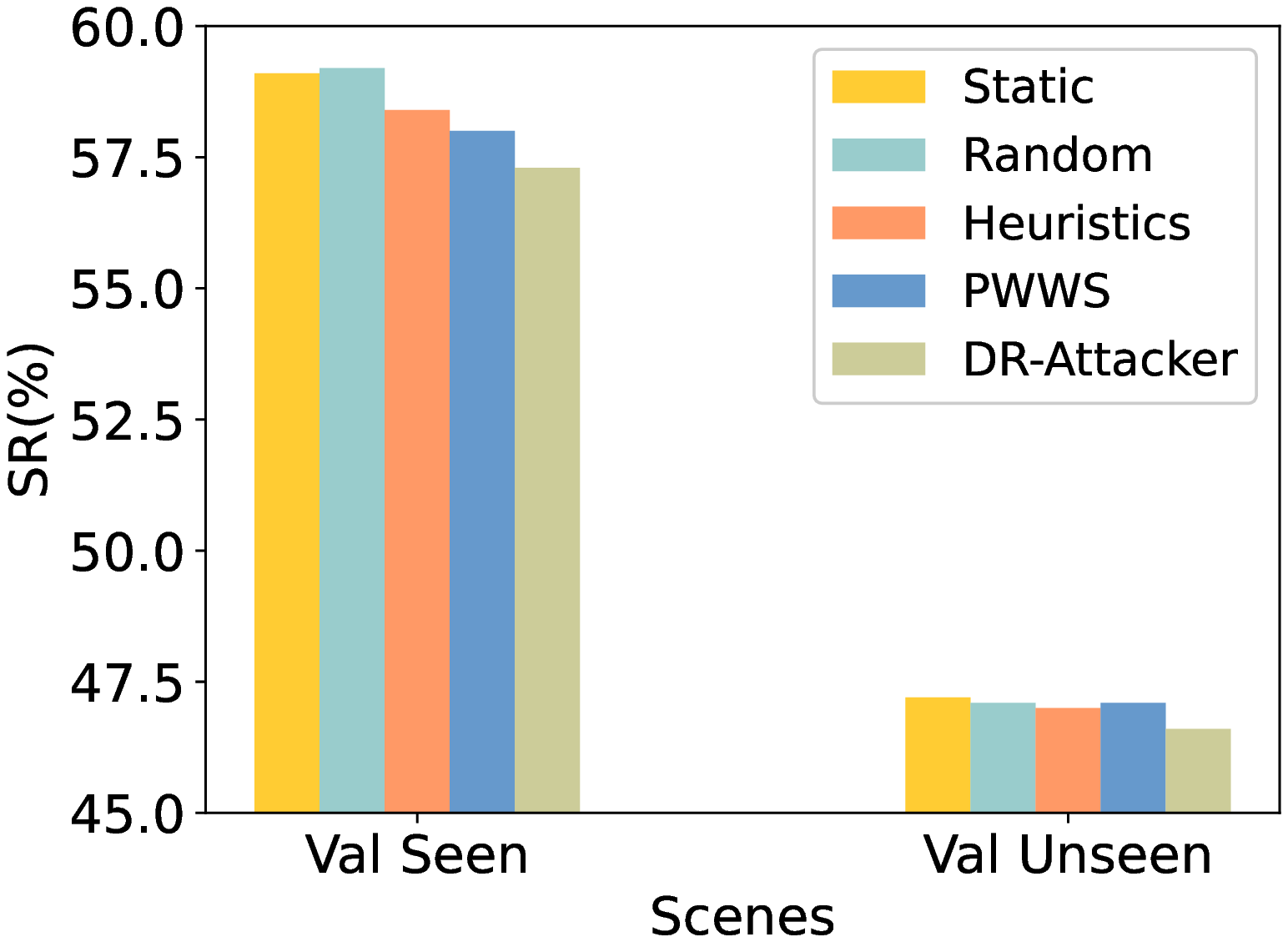}}
	\hspace{-0.01in}
	\subfloat[SPL(\%) for different scenes]{
		\includegraphics[width=5.5cm,height=4cm]{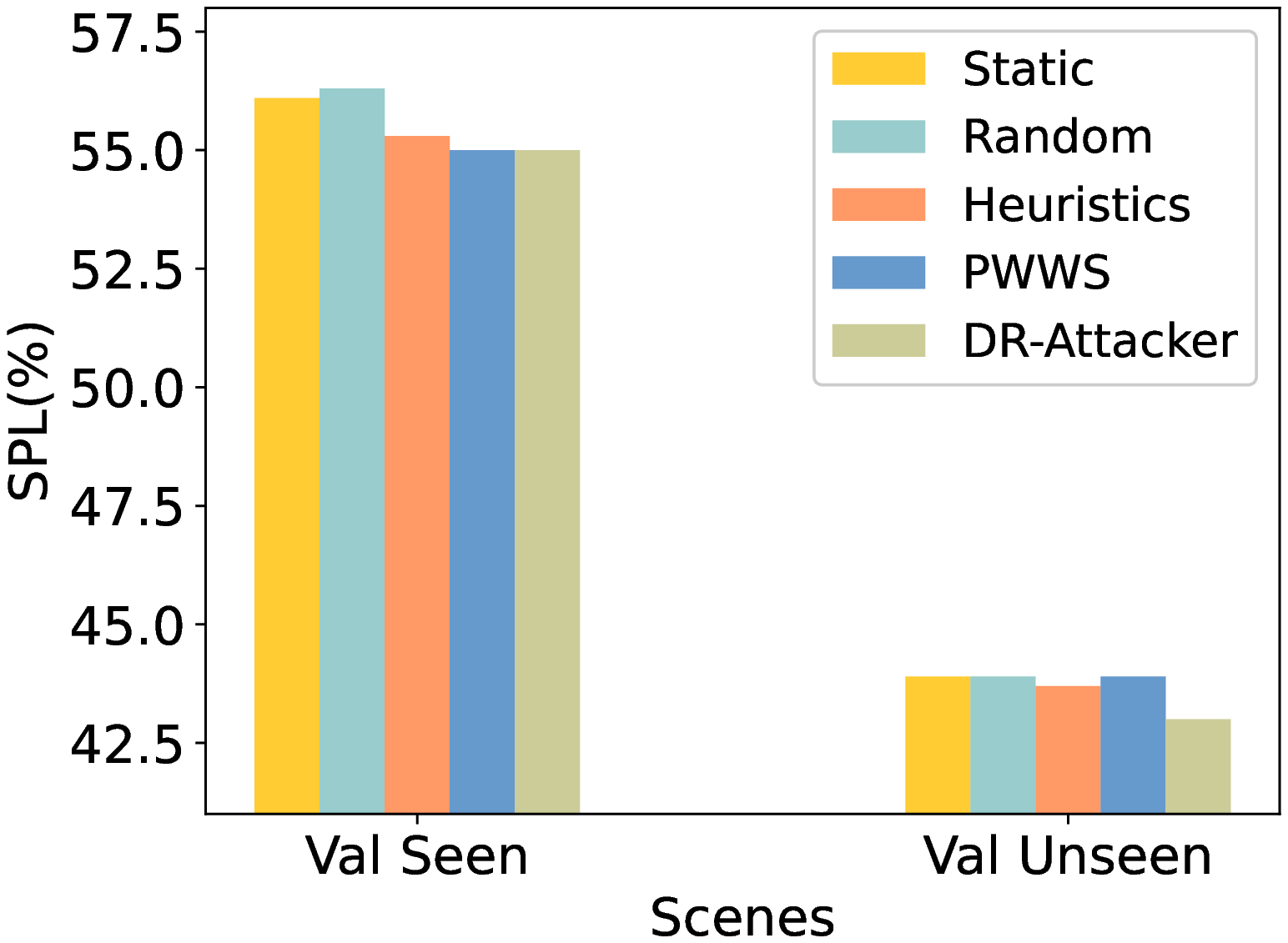}}
	\caption{The comparison results of different types of adversarial attacking mechanisms on VLN. NE (m), SR (\%) and SPL (\%) are reported for both Val Seen and Val Unseen scenes. Apart from NE, lower value indicates better results.}
		\vspace{-0.2cm}
	\label{tab:attack-vln} 
\end{figure*}

 \begin{figure*}
	\centering
	\includegraphics[width=1.0\linewidth]{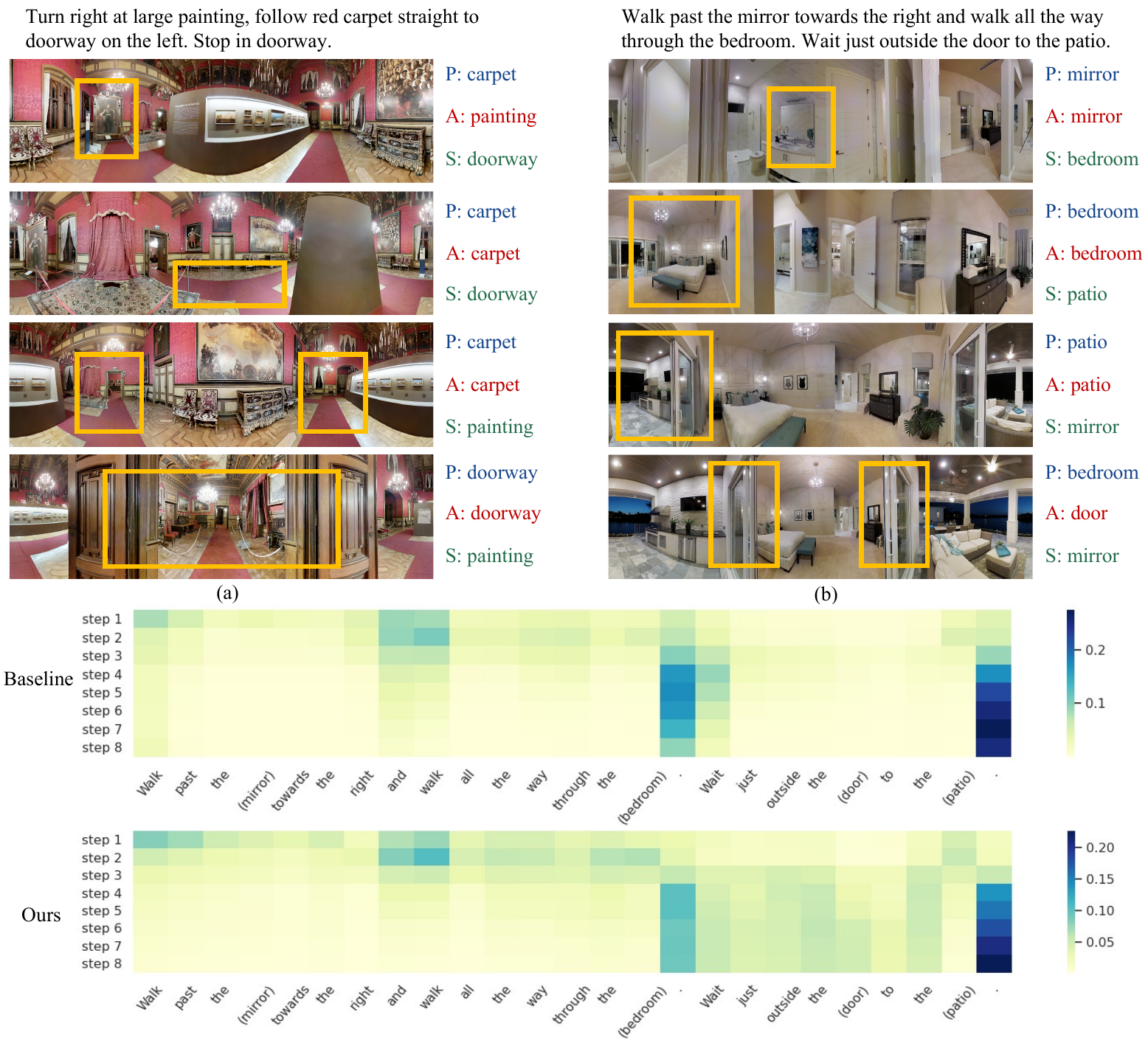}
	\caption{
		The visualization examples of perturbed instructions, panoramic views, and language attention weights (instance (b)) during trajectories on VLN. The words in red, blue and green color represent the actual attacked word by DR-Attacker (A), the predicted attacked word by the navigator (P),  and the substitution word (S),   respectively. Yellow bounding box denotes the  visual object or location at the current scene. ``Baseline'' and ``Ours'' represent the navigators trained without and with perturbed instructions, respectively. Words in the bracket represent the actual attacked word by the DR-Attacker. Best viewed in color.}
	
		\vspace{-0.4cm}
	\label{fig:visualization}

\end{figure*}

\begin{figure*}
	\centering
	\includegraphics[width=1.0\linewidth]{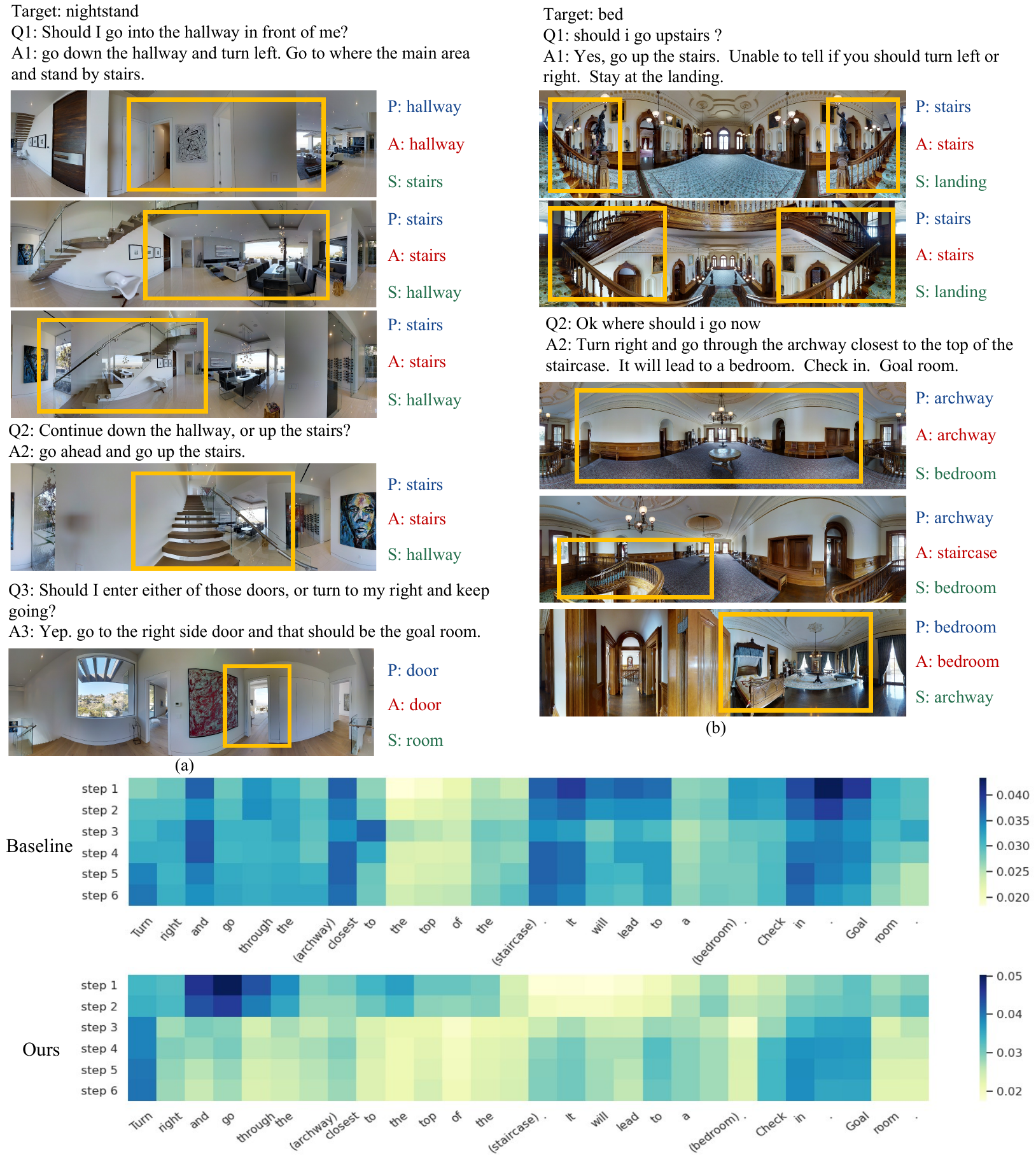}
	\caption{
		The visualization examples of perturbed instructions, panoramic views, and language attention weights (A2 in instance (b)) during trajectories on NDH. The words in red, blue and green color represent the actual attacked word by DR-Attacker (A), the predicted attacked word by the navigator (P),  and the substitution word (S),   respectively. Yellow bounding box denotes the  visual object or location at the current scene. ``Baseline'' and ``Ours'' represent the navigators trained without and with perturbed instructions, respectively. Words in the bracket represent the actual attacked word by the DR-Attacker. Best viewed in color.
	}
	\label{fig:NDH-visualization}
		\vspace{-0.4cm}
\end{figure*}

	\subsection{Quantitative Results}
	\label{Quantitative Results}

	\subsubsection{Comparison with the State-of-the-art Methods}

The quantitative comparison results with state-of-the-art methods on VLN and NDH are given in Table \ref{tab:comparison results on VLN} and Table \ref{tab:comparison results on NDH}, respectively. In Table \ref{tab:comparison results on VLN}, we report three most important metrics in the VLN setting, i.e., Navigation Error (NE), Success Rate (SR) and Success rate weighted by Path Length (SPL). In Table \ref{tab:comparison results on NDH}, we report the Goal Progress (GP) metric under the whole dialog history setting following most existing works on VDN \cite{thomason2019vision,zhu2020vision,hao2020towards}. 

Table \ref{tab:comparison results on VLN} indicates that our proposed method outperforms other competitors in most metrics. 
Comparing with the baseline EnvDrop \cite{tan2019learning}, the improvements for the SR and SPL of our method are significant in both seen and unseen settings. 
Table \ref{tab:comparison results on NDH} shows that our method outperforms the state-of-the-art methods by a significant margin on NDH in both seen and unseen environments. 
We further compare the training time, data and device between the state-of-the-art method PREVALENT \cite{hao2020towards} and our method on NDH. Since only the implementation of finetuning phase\footnote{https://github.com/weituo12321/PREVALENT\_R2R} is available for PREVALENT \cite{hao2020towards}, we only record the reimplemented finetuning time of PREVALENT \cite{hao2020towards} for comparison. Other values for the pretraining phase of PREVALENT \cite{hao2020towards} are the reported values in their paper. The results are given in Table \ref{tab:comparison-PREVALENT-ours}. From Table \ref{tab:comparison-PREVALENT-ours} we can find that compared with PREVALENT \cite{hao2020towards}, our proposed method need significantly less training time, data, and computation resource while can achieve better results, showing the good flexibility of our method.
Both the results on VLN and NDH show the effectiveness of the proposed method in improving the robustness of the navigation agent.

\subsubsection{Ablation Study}
In this section, we conduct ablation study to validate the effectiveness of the proposed adversarial attacking paradigm, adversarial training strategy and the auxiliary self-supervised reasoning task. Specifically, the effects of four-stage training for VLN and NDH tasks are presented in Table \ref{tab:ablation1} and Table \ref{tab:ablation2}. The effectiveness of the auxiliary self-supervised reasoning task is given in Table \ref{tab:DR-Attacker and Adversarial Training on NDH}. 
For VLN, ``Base Agent'' means pre-training navigators on the datasets composing of original instructions and augmented instructions for 40K iterations. ``Finetune'' means finetuning the adversarial trained agents on the same dataset as that used in the pretraining stage. For VDN, ``Base Agent'' means using the same training strategy like \cite{tan2019learning} to pre-train the navigators on the original dataset for 5k iterations. ``Finetune'' means finetuning the adversarial trained agents on the original dataset. ``DR-Attacker'' represents the navigation results when receiving perturbed instructions.
``Last A'', ``Last QA'' and ``All'' represent three kinds of different dialog history settings, i.e., the instruction is last answer, last question-answer pair or the whole dialog history \cite{thomason2019vision}.
    
From Table \ref{tab:ablation1} and Table \ref{tab:ablation2} we can find that our proposed four-stage training strategy can effectively contribute to enhancing the robustness and the navigation performance of the agent on both VLN and NDH tasks. Specifically, by introducing adversarial perturbations on the instructions, the navigation performance of the agent shows significant drop, demonstrating the effectiveness of the proposed adversarial attacking mechanism. Then, after adversarial training with the proposed auxiliary self-supervised reasoning task followed by finetuning on the original dataset, the robustness and the navigation performance can be effectively improved. Moreover, from Table  \ref{tab:DR-Attacker and Adversarial Training on NDH} we can observe that by introducing our proposed self-supervised auxiliary reasoning task in the adversarial training stage, the navigation performance can be effectively enhanced, demonstrating that improving the cross-modality understanding ability of the agent is crucial for successful navigation.  
 
\subsubsection{Different Types of Attacking Mechanisms}   
     In this subsection, we compare different types of attacking mechanisms to validate the effectiveness of the proposed DR-Attacker and in attacking and promoting the navigation performance through adversarial training. Specifically, four adversarial attacking methods or variants are chosen for the comparison: 1) ``Static'' means that the perturbation at each timestep is invariant, i.e., at each timestep,  the same target word is substituted with the same candidate word. For selecting the target word and candidate word, we use the pre-trained DR-Attacker to conduct the word prediction at the first navigation timestep. 2) ``Random'' represents randomly selecting the target word and the candidate substitution word at each timestep. 3)``Heuristics'' means the instruction word that receives the highest textual attention weights from the navigator at each timestep is destroyed. 4) PWWS \mbox{\cite{ren2019generating}} is an adversarial attack method in NLP which is similar to our proposed adversarial attack in some implementation procedures. It also obtains an attack score by calculating word importance and substitution impact according to the change of  classification probability. Since there is no direct classification-based objective for the instruction in both VLN and NDH tasks, we choose the action prediction  probability for an alternative. Specifically, at each timestep, the attacked word which can cause the maximum change of the original action prediction  probability is destroyed. Therefore, ``Random'', ``Heursitics'' and PWWS are all dynamic adversarial attacks.
    
    The comparison results of attacking effects on VLN and NDH tasks are given in Figure \ref{tab:attack-vln} and Table \ref{tab:adv on NDH}, respectively. And the adversarial training results using different attacking mechanisms on NDH are given in Table \ref{tab:adv on NDH}. From Figure \ref{tab:attack-vln} and Table \ref{tab:adv on NDH} we can find that compared with either static or dynamic attacking mechanisms, our proposed DR-Attacker can achieve the best attack results in most metrics on both VLN and NDH tasks, demonstrating the importance of dynamically attacking key information in the navigation task and the effectiveness of our proposed RL-based optimization method for the proposed adversarial attack. Moreover, from the adversarial training results in Table \ref{tab:adv on NDH} we can find the superiority of DR-Attacker in promoting the navigation performance compared with other attacking methods, demonstrating that jointly optimizing the navigator and the attacker is more beneficial for the improvement of the navigation performance. Both the attacking and adversarial training results on VLN and NDH tasks show the effectiveness of the proposed adversarial attacking mechanism and adversarial training paradigm.

    \subsection{Qualitative Results}
    \label{Qualitative Results}
    In this subsection, we show the visualization examples of perturbed instructions, panoramic views and language attention weights during trajectories on VLN and NDH tasks. The results are given in Figure \ref{fig:visualization} and Figure \ref{fig:NDH-visualization}, respectively.
    From Figure \ref{fig:visualization} and Figure \ref{fig:NDH-visualization} we can find that the proposed DR-Attacker can successfully locate the word which appears in the scene at different timesteps and substitute it with the word that doesn't exist in the current scene. Moreover, the navigator can make correct predictions of the actual attacked words by DR-Attacker, showing its good understanding of the multi-modality observations. The first subfigure in Figure \ref{fig:visualization} (a), the fourth subfigure in Figure \ref{fig:visualization} (b) and the second subfigure in Figure \ref{fig:NDH-visualization} (a) show the failure cases.
    From the failure cases, we can find that when there are multiple objects referred in the instruction   simultaneously existing in the current scene, e.g., both the ``bedroom'' and ``door'' exist in the fourth subfigure in  Figure \ref{fig:visualization} (b), the navigator or the DR-Attacker may be confused. 
    From the language attention weights of the navigators trained with perturbed instructions (``Ours''), we can find that although the target word is attacked, the navigator can attend to the context near the attacked word to capture the language intention. Moreover, with the process of the navigation trajectory, it can successfully capture important  instruction information in different phases. In contrast, the navigator trained without perturbed instructions (``Baseline'') generates a confused language attention weights by the introduced perturbations during navigation.
    These visualization analyses show that emphasizing useful instruction information can contribute to successful navigation. Moreover, our proposed adversarial attacking and adversarial training mechanisms can effectively improve the robustness of the navigation agent. 

 \section{Conclusion}
    \label{conclusion}
    In this work, we propose Dynamic Reinforced Instruction Attacker (DR-Attacker) for the natural language grounded visual navigation  tasks. By formulating the perturbation generation using the RL framework, DR-Attacker can be optimized iteratively to capture the crucial parts in instructions and generate meaningful adversarial samples. Through adversarial training using perturbed instructions, the robustness of the navigator can be effectively enhanced with an auxiliary self-supervised reasoning task. Experiments on both VLN and NDH tasks show the effectiveness of the proposed method. 
    
    In the future, we plan to improve the training strategy of the proposed instruction attacker and exploit to design more effective attacks on the navigation instruction. Moreover, we would like to develop multi-modality adversarial attacks for the embodied navigation task to further verify and improve the robustness of the navigator. 
	
	
	%

	

	\section*{Acknowledgment}


This work was supported in part by National Key R\&D Program of China under Grant No. 2020AAA0109700, National Natural Science Foundation of China (NSFC) under Grant No.U19A2073 and No.61976233, Guangdong Province Basic and Applied Basic Research (Regional Joint Fund-Key) Grant No.2019B1515120039, Guangdong Outstanding Youth Fund (Grant No. 2021B1515020061), Shenzhen Fundamental Research Program (Project No. RCYX20200714114642083, No. JCYJ20190807154211365), Zhejiang Lab’s Open Fund (No. 2020AA3AB14) and CSIG Young Fellow Support Fund.

	\ifCLASSOPTIONcaptionsoff
	\newpage
	\fi

	
	
	%
		
		
\vspace{-0.2cm}
\bibliographystyle{IEEEtran}
\bibliography{IEEEabrv,egbib}	

\begin{thebibliography}{10}
\providecommand{\url}[1]{#1}
\csname url@samestyle\endcsname
\providecommand{\newblock}{\relax}
\providecommand{\bibinfo}[2]{#2}
\providecommand{\BIBentrySTDinterwordspacing}{\spaceskip=0pt\relax}
\providecommand{\BIBentryALTinterwordstretchfactor}{4}
\providecommand{\BIBentryALTinterwordspacing}{\spaceskip=\fontdimen2\font plus
\BIBentryALTinterwordstretchfactor\fontdimen3\font minus
  \fontdimen4\font\relax}
\providecommand{\BIBforeignlanguage}[2]{{%
\expandafter\ifx\csname l@#1\endcsname\relax
\typeout{** WARNING: IEEEtran.bst: No hyphenation pattern has been}%
\typeout{** loaded for the language `#1'. Using the pattern for}%
\typeout{** the default language instead.}%
\else
\language=\csname l@#1\endcsname
\fi
#2}}
\providecommand{\BIBdecl}{\relax}
\BIBdecl

\bibitem{anderson2018vision}
P.~{Anderson}, Q.~{Wu}, D.~{Teney}, J.~{Bruce}, M.~{Johnson}, N.~{Sunderhauf},
  I.~{Reid}, S.~{Gould}, and A.~van~den {Hengel}, ``Vision-and-language
  navigation: Interpreting visually-grounded navigation instructions in real
  environments,'' in \emph{2018 IEEE/CVF Conference on Computer Vision and
  Pattern Recognition}, 2018, pp. 3674--3683.

\bibitem{thomason2019vision}
J.~{Thomason}, M.~{Murray}, M.~{Cakmak}, and L.~{Zettlemoyer},
  ``Vision-and-dialog navigation,'' \emph{Conference on Robot Learning (CoRL)},
  pp. 394--406, 2019.

\bibitem{qi2020reverie}
Y.~{Qi}, Q.~{Wu}, P.~{Anderson}, X.~{Wang}, W.~Y. {Wang}, C.~{Shen}, and
  A.~van~den {Hengel}, ``Reverie: Remote embodied visual referring expression
  in real indoor environments,'' in \emph{2020 IEEE/CVF Conference on Computer
  Vision and Pattern Recognition (CVPR)}, 2020, pp. 9982--9991.

\bibitem{nguyen2019help}
K.~{Nguyen} and H.~{Daumé}, ``Help, anna! vision-based navigation with natural
  multimodal assistance via retrospective curiosity-encouraging imitation
  learning,'' in \emph{2019 Conference on Empirical Methods in Natural Language
  Processing}, 2019, pp. 684--695.

\bibitem{chen2019touchdown}
H.~{Chen}, A.~{Suhr}, D.~{Misra}, N.~{Snavely}, and Y.~{Artzi}, ``Touchdown:
  Natural language navigation and spatial reasoning in visual street
  environments,'' in \emph{2019 IEEE/CVF Conference on Computer Vision and
  Pattern Recognition (CVPR)}, 2019, pp. 12\,538--12\,547.

\bibitem{fried2018speaker}
D.~{Fried}, R.~{Hu}, V.~{Cirik}, A.~{Rohrbach}, J.~{Andreas}, L.-P. {Morency},
  T.~{Berg-Kirkpatrick}, K.~{Saenko}, D.~{Klein}, and T.~{Darrell},
  ``Speaker-follower models for vision-and-language navigation,'' in \emph{NIPS
  2018: The 32nd Annual Conference on Neural Information Processing Systems},
  2018, pp. 3314--3325.

\bibitem{tan2019learning}
H.~{Tan}, L.~{Yu}, and M.~{Bansal}, ``Learning to navigate unseen environments:
  Back translation with environmental dropout,'' in \emph{NAACL-HLT 2019:
  Annual Conference of the North American Chapter of the Association for
  Computational Linguistics}, 2019, pp. 2610--2621.

\bibitem{fu2020counterfactual}
T.-J. {Fu}, X.~E. {Wang}, M.~F. {Peterson}, S.~T. {Grafton}, M.~P. {Eckstein},
  and W.~Y. {Wang}, ``Counterfactual vision-and-language navigation via
  adversarial path sampler.'' in \emph{European Conference on Computer Vision},
  2020, pp. 71--86.

\bibitem{araujo2020on}
V.~{Araujo}, A.~{Carvallo}, C.~{Aspillaga}, and D.~{Parra}, ``On adversarial
  examples for biomedical nlp tasks,'' \emph{arXiv preprint arXiv:2004.11157},
  2020.

\bibitem{li2020bert}
L.~{Li}, R.~{Ma}, Q.~{Guo}, X.~{Xue}, and X.~{Qiu}, ``Bert-attack: Adversarial
  attack against bert using bert,'' in \emph{Proceedings of the 2020 Conference
  on Empirical Methods in Natural Language Processing (EMNLP)}, 2020, pp.
  6193--6202.

\bibitem{ebrahimi2018on}
J.~{Ebrahimi}, D.~{Lowd}, and D.~{Dou}, ``On adversarial examples for
  character-level neural machine translation,'' in \emph{COLING 2018: 27th
  International Conference on Computational Linguistics}, 2018, pp. 653--663.

\bibitem{ren2019generating}
S.~{Ren}, Y.~{Deng}, K.~{He}, and W.~{Che}, ``Generating natural language
  adversarial examples through probability weighted word saliency,'' in
  \emph{ACL 2019 : The 57th Annual Meeting of the Association for Computational
  Linguistics}, 2019, pp. 1085--1097.

\bibitem{zang2020word}
Y.~{Zang}, F.~{Qi}, C.~{Yang}, Z.~{Liu}, M.~{Zhang}, Q.~{Liu}, and M.~{Sun},
  ``Word-level textual adversarial attacking as combinatorial optimization,''
  in \emph{Proceedings of the 58th Annual Meeting of the Association for
  Computational Linguistics}, 2020, pp. 6066--6080.

\bibitem{wang2020environment}
X.~E. {Wang}, V.~{Jain}, E.~{Ie}, W.~Y. {Wang}, Z.~{Kozareva}, and S.~{Ravi},
  ``Environment-agnostic multitask learning for natural language grounded
  navigation,'' in \emph{ECCV (24)}, 2020, pp. 413--430.

\bibitem{nguyen2019vision}
K.~{Nguyen}, D.~{Dey}, C.~{Brockett}, and B.~{Dolan}, ``Vision-based navigation
  with language-based assistance via imitation learning with indirect
  intervention,'' in \emph{2019 IEEE/CVF Conference on Computer Vision and
  Pattern Recognition (CVPR)}, 2019, pp. 12\,527--12\,537.

\bibitem{antol2015vqa}
S.~{Antol}, A.~{Agrawal}, J.~{Lu}, M.~{Mitchell}, D.~{Batra}, C.~L. {Zitnick},
  and D.~{Parikh}, ``Vqa: Visual question answering,'' in \emph{2015 IEEE
  International Conference on Computer Vision (ICCV)}, 2015, pp. 2425--2433.

\bibitem{vries2017guesswhat}
\BIBentryALTinterwordspacing
H.~de~{Vries}, F.~{Strub}, S.~{Chandar}, O.~{Pietquin}, H.~{Larochelle}, and
  A.~{Courville}, ``Guesswhat?! visual object discovery through multi-modal
  dialogue,'' in \emph{2017 IEEE Conference on Computer Vision and Pattern
  Recognition (CVPR)}, 2017, pp. 4466--4475. [Online]. Available:
  \url{https://academic.microsoft.com/paper/2558809543}
\BIBentrySTDinterwordspacing

\bibitem{das2017visual}
\BIBentryALTinterwordspacing
A.~{Das}, S.~{Kottur}, K.~{Gupta}, A.~{Singh}, D.~{Yadav}, J.~M.~F. {Moura},
  D.~{Parikh}, and D.~{Batra}, ``Visual dialog,'' in \emph{2017 IEEE Conference
  on Computer Vision and Pattern Recognition (CVPR)}, 2017. [Online].
  Available: \url{https://academic.microsoft.com/paper/2768661419}
\BIBentrySTDinterwordspacing

\bibitem{you2016image}
Q.~{You}, H.~{Jin}, Z.~{Wang}, C.~{Fang}, and J.~{Luo}, ``Image captioning with
  semantic attention,'' in \emph{2016 IEEE Conference on Computer Vision and
  Pattern Recognition (CVPR)}, 2016, pp. 4651--4659.

\bibitem{wang2019reinforced}
X.~{Wang}, Q.~{Huang}, A.~{Celikyilmaz}, J.~{Gao}, D.~{Shen}, Y.-F. {Wang},
  W.~Y. {Wang}, and L.~{Zhang}, ``Reinforced cross-modal matching and
  self-supervised imitation learning for vision-language navigation,'' in
  \emph{2019 IEEE/CVF Conference on Computer Vision and Pattern Recognition
  (CVPR)}, 2019, pp. 6629--6638.

\bibitem{ma2019self}
C.-Y. {Ma}, jiasen {lu}, Z.~{Wu}, G.~{AlRegib}, Z.~{Kira}, richard {socher},
  and C.~{Xiong}, ``Self-monitoring navigation agent via auxiliary progress
  estimation,'' in \emph{ICLR 2019 : 7th International Conference on Learning
  Representations}, 2019.

\bibitem{zhu2020vision}
Y.~{Zhu}, F.~{Zhu}, Z.~{Zhan}, B.~{Lin}, J.~{Jiao}, X.~{Chang}, and X.~{Liang},
  ``Vision-dialog navigation by exploring cross-modal memory,'' in \emph{2020
  IEEE/CVF Conference on Computer Vision and Pattern Recognition (CVPR)}, 2020,
  pp. 10\,730--10\,739.

\bibitem{hao2020towards}
W.~{Hao}, C.~{Li}, X.~{Li}, L.~{Carin}, and J.~{Gao}, ``Towards learning a
  generic agent for vision-and-language navigation via pre-training,'' in
  \emph{2020 IEEE/CVF Conference on Computer Vision and Pattern Recognition
  (CVPR)}, 2020, pp. 13\,137--13\,146.

\bibitem{li2019robust}
X.~{Li}, C.~{Li}, Q.~{Xia}, Y.~{Bisk}, A.~{Çelikyilmaz}, J.~{Gao}, N.~A.
  {Smith}, and Y.~{Choi}, ``Robust navigation with language pretraining and
  stochastic sampling.'' in \emph{Proceedings of the 2019 Conference on
  Empirical Methods in Natural Language Processing and the 9th International
  Joint Conference on Natural Language Processing (EMNLP-IJCNLP)}, 2019, pp.
  1494--1499.

\bibitem{cemgil2020adversarially}
T.~{Cemgil}, S.~{Ghaisas}, K.~{Dvijotham}, and P.~{Kohli}, ``Adversarially
  robust representations with smooth encoders,'' in \emph{ICLR 2020 : Eighth
  International Conference on Learning Representations}, 2020.

\bibitem{zhang2019defense}
H.~{Zhang} and J.~{Wang}, ``Defense against adversarial attacks using feature
  scattering-based adversarial training,'' in \emph{NeurIPS 2019 : Thirty-third
  Conference on Neural Information Processing Systems}, 2019, pp. 1831--1841.

\bibitem{salman2019provably}
H.~{Salman}, J.~{Li}, I.~{Razenshteyn}, P.~{Zhang}, H.~{Zhang}, S.~{Bubeck},
  and G.~{Yang}, ``Provably robust deep learning via adversarially trained
  smoothed classifiers,'' in \emph{NeurIPS 2019 : Thirty-third Conference on
  Neural Information Processing Systems}, 2019, pp. 11\,292--11\,303.

\bibitem{feng2019learning}
J.~{Feng}, Q.-Z. {Cai}, and Z.-H. {Zhou}, ``Learning to confuse: Generating
  training time adversarial data with auto-encoder,'' in \emph{NeurIPS 2019 :
  Thirty-third Conference on Neural Information Processing Systems}, 2019, pp.
  11\,994--12\,004.

\bibitem{mopuri2019generalizable}
K.~R. {Mopuri}, A.~{Ganeshan}, and R.~V. {Babu}, ``Generalizable data-free
  objective for crafting universal adversarial perturbations,'' \emph{IEEE
  Transactions on Pattern Analysis and Machine Intelligence}, vol.~41, no.~10,
  pp. 2452--2465, 2019.

\bibitem{he2016deep}
K.~{He}, X.~{Zhang}, S.~{Ren}, and J.~{Sun}, ``Deep residual learning for image
  recognition,'' in \emph{2016 IEEE Conference on Computer Vision and Pattern
  Recognition (CVPR)}, 2016, pp. 770--778.

\bibitem{simonyan2015very}
K.~{Simonyan} and A.~{Zisserman}, ``Very deep convolutional networks for
  large-scale image recognition,'' in \emph{ICLR 2015 : International
  Conference on Learning Representations 2015}, 2015.

\bibitem{wang2019improving}
D.~{Wang}, C.~{Gong}, and Q.~{Liu}, ``Improving neural language modeling via
  adversarial training,'' in \emph{ICML 2019 : Thirty-sixth International
  Conference on Machine Learning}, 2019, pp. 6555--6565.

\bibitem{zhu2019freelb}
C.~{Zhu}, Y.~{Cheng}, Z.~{Gan}, S.~{Sun}, T.~{Goldstein}, and J.~{Liu},
  ``Freelb: Enhanced adversarial training for natural language understanding,''
  in \emph{International Conference on Learning Representations}, 2019.

\bibitem{cheng2019robust}
Y.~{Cheng}, L.~{Jiang}, and W.~{Macherey}, ``Robust neural machine translation
  with doubly adversarial inputs,'' in \emph{ACL 2019 : The 57th Annual Meeting
  of the Association for Computational Linguistics}, 2019, pp. 4324--4333.

\bibitem{jones2020robust}
E.~{Jones}, R.~{Jia}, A.~{Raghunathan}, and P.~{Liang}, ``Robust encodings: A
  framework for combating adversarial typos.'' in \emph{Proceedings of the 58th
  Annual Meeting of the Association for Computational Linguistics}, 2020, pp.
  2752--2765.

\bibitem{ebrahimi2018hotflip}
J.~{Ebrahimi}, A.~{Rao}, D.~{Lowd}, and D.~{Dou}, ``Hotflip: White-box
  adversarial examples for text classification,'' in \emph{ACL 2018: 56th
  Annual Meeting of the Association for Computational Linguistics}, vol.~2,
  2018, pp. 31--36.

\bibitem{conneau2017supervised}
A.~{Conneau}, D.~{Kiela}, H.~{Schwenk}, L.~{Barrault}, and A.~{Bordes},
  ``Supervised learning of universal sentence representations from natural
  language inference data,'' in \emph{Proceedings of the 2017 Conference on
  Empirical Methods in Natural Language Processing}, 2017, pp. 670--680.

\bibitem{devlin2019bert}
J.~{Devlin}, M.-W. {Chang}, K.~{Lee}, and K.~{Toutanova}, ``Bert: Pre-training
  of deep bidirectional transformers for language understanding,'' in
  \emph{NAACL-HLT 2019: Annual Conference of the North American Chapter of the
  Association for Computational Linguistics}, 2019, pp. 4171--4186.

\bibitem{wang2021infobert}
B.~{Wang}, S.~{Wang}, Y.~{Cheng}, Z.~{Gan}, R.~{Jia}, B.~{Li}, and J.~{Liu},
  ``Infobert: Improving robustness of language models from an information
  theoretic perspective,'' in \emph{ICLR 2021: The Ninth International
  Conference on Learning Representations}, 2021.

\bibitem{jia2019certified}
R.~{Jia}, A.~{Raghunathan}, K.~{Göksel}, and P.~{Liang}, ``Certified
  robustness to adversarial word substitutions,'' in \emph{2019 Conference on
  Empirical Methods in Natural Language Processing}, 2019, pp. 4127--4140.

\bibitem{eger2019text}
S.~{Eger}, G.~G. {Sahin}, A.~{Rücklé}, J.-U. {Lee}, C.~{Schulz}, M.~{Mesgar},
  K.~{Swarnkar}, E.~{Simpson}, and I.~{Gurevych}, ``Text processing like humans
  do: Visually attacking and shielding nlp systems,'' in \emph{NAACL-HLT 2019:
  Annual Conference of the North American Chapter of the Association for
  Computational Linguistics}, 2019, pp. 1634--1647.

\bibitem{liu2020adversarial}
X.~{Liu}, H.~{Cheng}, P.~{He}, W.~{Chen}, Y.~{Wang}, H.~{Poon}, and J.~{Gao},
  ``Adversarial training for large neural language models,'' \emph{arXiv
  preprint arXiv:2004.08994}, 2020.

\bibitem{yin2020on}
F.~{Yin}, Q.~{Long}, T.~{Meng}, and K.-W. {Chang}, ``On the robustness of
  language encoders against grammatical errors,'' in \emph{Proceedings of the
  58th Annual Meeting of the Association for Computational Linguistics}, 2020,
  pp. 3386--3403.

\bibitem{liu2020spatiotemporal}
A.~{Liu}, T.~{Huang}, X.~{Liu}, Y.~{Xu}, Y.~{Ma}, X.~{Chen}, S.~J. {Maybank},
  and D.~{Tao}, ``Spatiotemporal attacks for embodied agents.'' in \emph{ECCV
  (17)}, 2020, pp. 122--138.

\bibitem{das2018embodied}
A.~{Das}, S.~{Datta}, G.~{Gkioxari}, S.~{Lee}, D.~{Parikh}, and D.~{Batra},
  ``Embodied question answering,'' in \emph{2018 IEEE/CVF Conference on
  Computer Vision and Pattern Recognition Workshops (CVPRW)}, 2018, pp. 1--10.

\bibitem{cubuk2019autoaugment}
E.~D. {Cubuk}, B.~{Zoph}, D.~{Mane}, V.~{Vasudevan}, and Q.~V. {Le},
  ``Autoaugment: Learning augmentation strategies from data,'' in \emph{2019
  IEEE/CVF Conference on Computer Vision and Pattern Recognition (CVPR)}, 2019,
  pp. 113--123.

\bibitem{ho2019population}
D.~{Ho}, E.~{Liang}, X.~{Chen}, I.~{Stoica}, and P.~{Abbeel}, ``Population
  based augmentation: Efficient learning of augmentation policy schedules,'' in
  \emph{International Conference on Machine Learning}, 2019, pp. 2731--2741.

\bibitem{lim2019fast}
S.~{Lim}, I.~{Kim}, T.~{Kim}, C.~{Kim}, and S.~{Kim}, ``Fast autoaugment,'' in
  \emph{Advances in Neural Information Processing Systems}, vol.~32, 2019, pp.
  6665--6675.

\bibitem{hataya2020faster}
R.~{Hataya}, J.~{Zdenek}, K.~{Yoshizoe}, and H.~{Nakayama}, ``Faster
  autoaugment: Learning augmentation strategies using backpropagation.'' in
  \emph{ECCV (25)}, 2020, pp. 1--16.

\bibitem{cubuk2020randaugment}
E.~D. {Cubuk}, B.~{Zoph}, J.~{Shlens}, and Q.~{Le}, ``Randaugment: Practical
  automated data augmentation with a reduced search space,'' in \emph{Advances
  in Neural Information Processing Systems}, vol.~33, 2020, pp.
  18\,613--18\,624.

\bibitem{wu2018building}
Y.~{Wu}, Y.~{Wu}, G.~{Gkioxari}, and Y.~{Tian}, ``Building generalizable agents
  with a realistic and rich 3d environment,'' in \emph{ICLR 2018 :
  International Conference on Learning Representations 2018}, 2018.

\bibitem{mnih2016asynchronous}
V.~{Mnih}, A.~P. {Badia}, M.~{Mirza}, A.~{Graves}, T.~{Harley}, T.~P.
  {Lillicrap}, D.~{Silver}, and K.~{Kavukcuoglu}, ``Asynchronous methods for
  deep reinforcement learning,'' in \emph{ICML'16 Proceedings of the 33rd
  International Conference on International Conference on Machine Learning -
  Volume 48}, 2016, pp. 1928--1937.

\bibitem{pinto2017robust}
L.~{Pinto}, J.~{Davidson}, R.~{Sukthankar}, and A.~{Gupta}, ``Robust
  adversarial reinforcement learning,'' in \emph{ICML'17 Proceedings of the
  34th International Conference on Machine Learning - Volume 70}, 2017, pp.
  2817--2826.

\bibitem{ma2019the}
C.-Y. {Ma}, Z.~{Wu}, G.~{AlRegib}, C.~{Xiong}, and Z.~{Kira}, ``The regretful
  agent: Heuristic-aided navigation through progress estimation,'' in
  \emph{2019 IEEE/CVF Conference on Computer Vision and Pattern Recognition
  (CVPR)}, 2019, pp. 6732--6740.

\end{thebibliography}
	
	%

\begin{IEEEbiography}[{\includegraphics[width=1in,height=1.25in,clip,keepaspectratio]{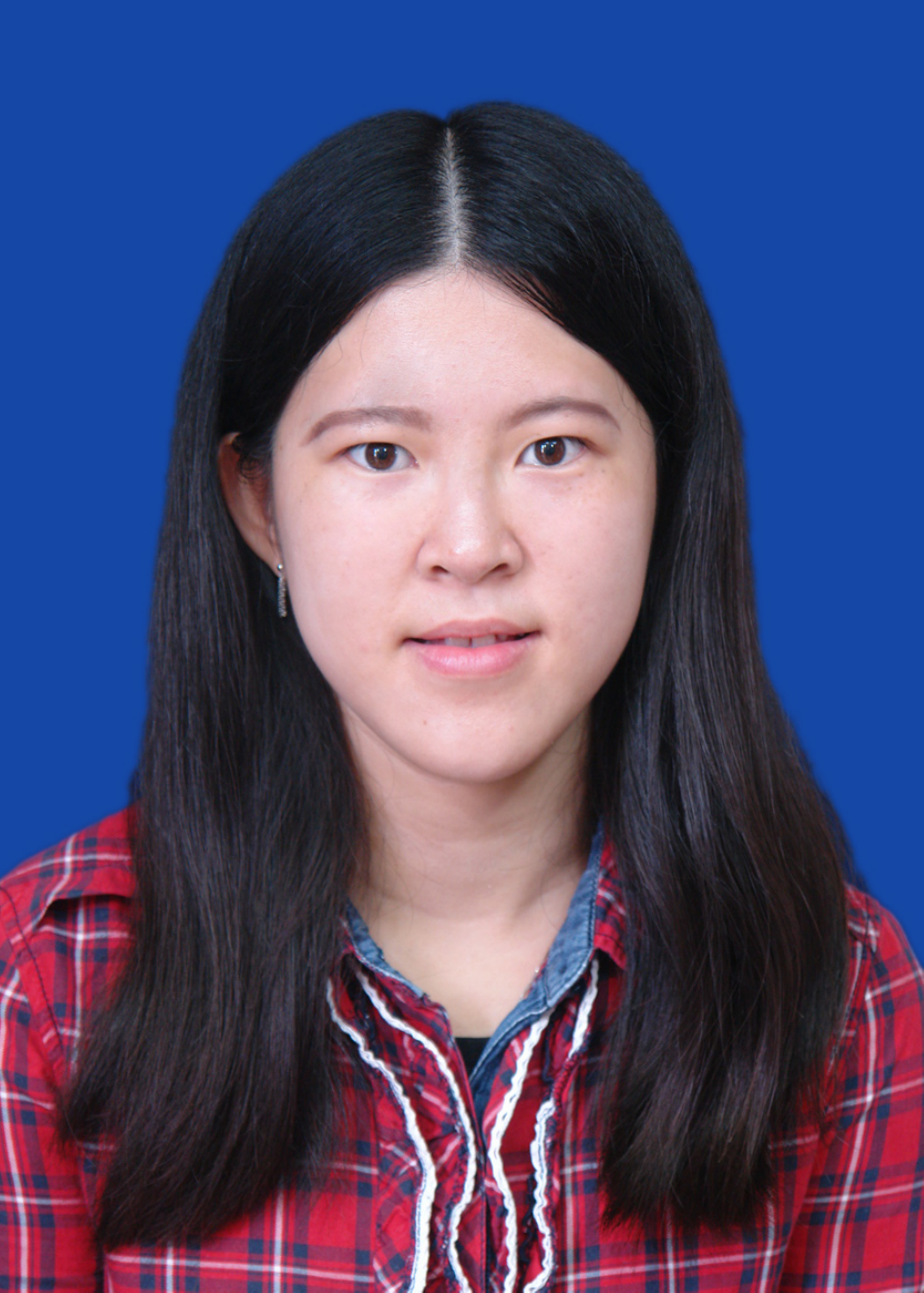}}]{Bingqian Lin} received the B.E. and the M.E.
degree in Computer Science from University of
Electronic Science and Technology of China and
Xiamen University, in 2016 and 2019, respectively.
She is currently working toward the D.Eng in the
school of intelligent systems engineering of Sun Yat-sen University. Her research interests include multi-view clustering, image processing and vision-and-language understanding.
\end{IEEEbiography}

\begin{IEEEbiography}[{\includegraphics[width=1in,height=1.25in,clip,keepaspectratio]{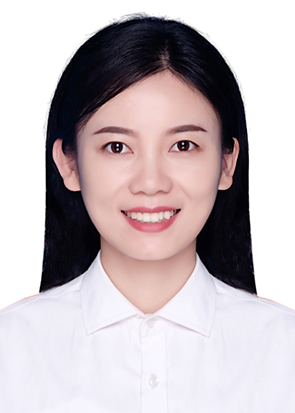}}]{Yi Zhu} received the B.S. degree in software engineering from Sun Yat-sen University, Guangzhou, China, in 2013. Since 2015, she has been a Ph.D student in computer science with the School of Electronic, Electrical, and Communication Engineering, University of Chinese Academy of Sciences, Beijing, China. Her current research interests include object recognition, scene understanding, weakly supervised learning and visual reasoning.
\end{IEEEbiography}

\begin{IEEEbiography}[{\includegraphics[width=1in,height=1.25in,clip,keepaspectratio]{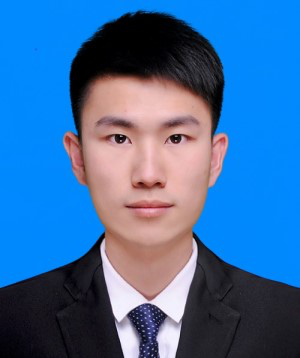}}]{Yanxin Long} is a first-year master in the School of Intelligent Systems Engineering, Sun Yat-Sen University. He works at the Human Cyber Physical Intelligence Integration Lab under the supervision of Prof. Xiaodan Liang. Before that, He received my Bachelor Degree from the Comunication College, Xidian University in 2020. His research interests  include reinforcement learning and vision-and-language understanding.
\end{IEEEbiography}

\begin{IEEEbiography}[{\includegraphics[width=1in,height=1.25in,clip,keepaspectratio]{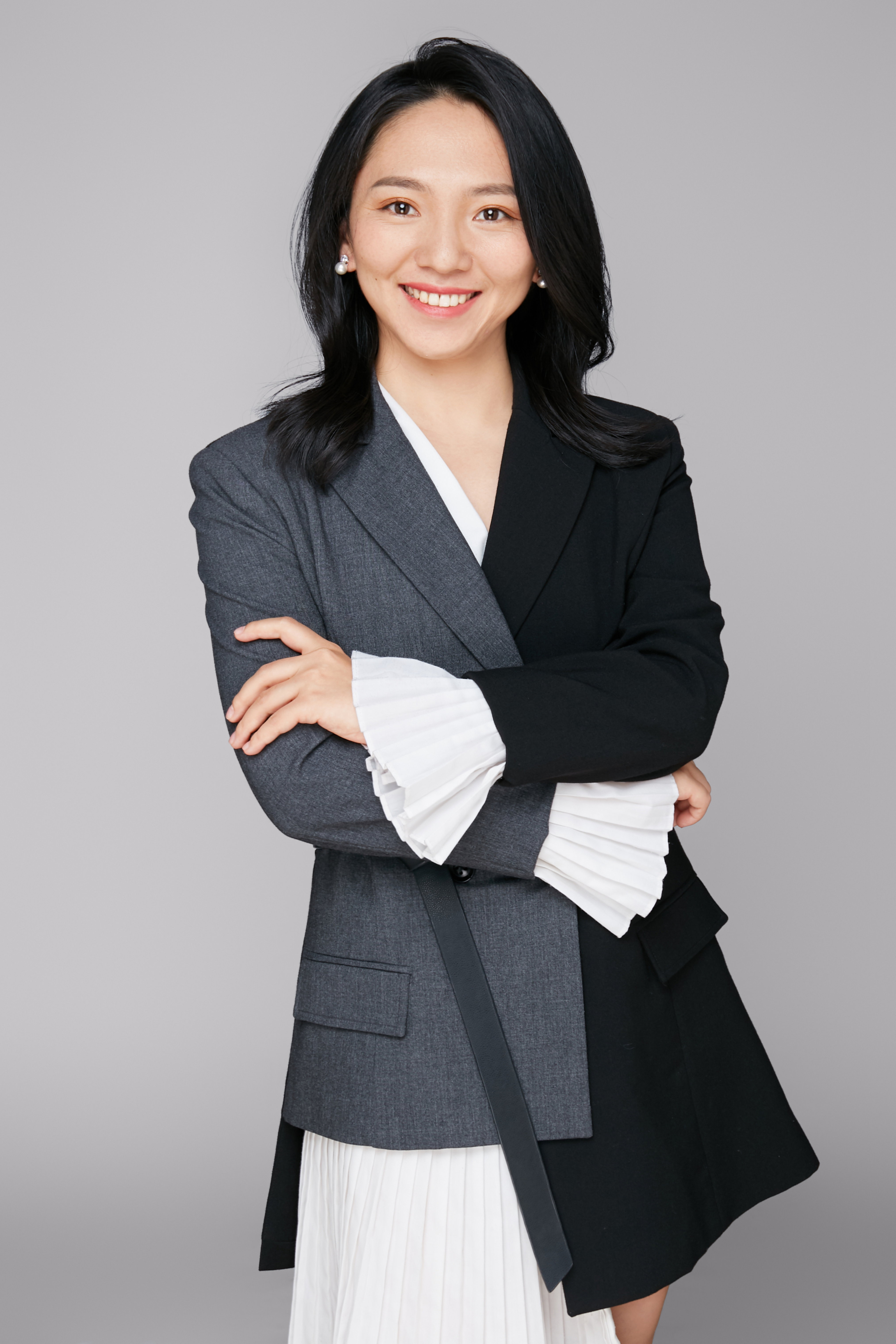}}]{Xiaodan Liang} is currently an Associate Professor at Sun Yat-sen University. She was a postdoc researcher in the machine learning department at Carnegie Mellon University, working with Prof. Eric Xing, from 2016 to 2018. She received her PhD degree from Sun Yat-sen University in 2016, advised by Liang Lin. She has published several cutting-edge projects on human-related analysis, including human parsing, pedestrian detection and instance segmentation, 2D/3D human pose estimation and activity recognition.
\end{IEEEbiography}

\begin{IEEEbiography}[{\includegraphics[width=1in,height=1.25in,clip,keepaspectratio]{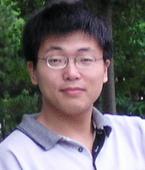}}]{Qixiang Ye} received the B.S. and M.S. degrees
in mechanical and electrical engineering from the
Harbin Institute of Technology, China, in 1999 and
2001, respectively, and the Ph.D. degree from the Institute
of Computing Technology, Chinese Academy
of Sciences, in 2006. He has been a Professor with
the University of Chinese Academy of Sciences
since 2009, and was a Visiting Assistant Professor
with the Institute of Advanced Computer Studies,
University of Maryland, College Park, in 2013. He
has authored over 50 papers in refereed conferences
and journals, and received the Sony Outstanding Paper Award. His current
research interests include image processing, visual object detection and
machine learning. He pioneered the Kernel SVM-based pyrolysis output
prediction software which was put into practical application by SINOPEC
in 2012. He developed two kinds of piecewise linear SVM methods which
were successfully applied into visual object detection.
\end{IEEEbiography}

\begin{IEEEbiography}[{\includegraphics[width=1in,height=1.25in,clip,keepaspectratio]{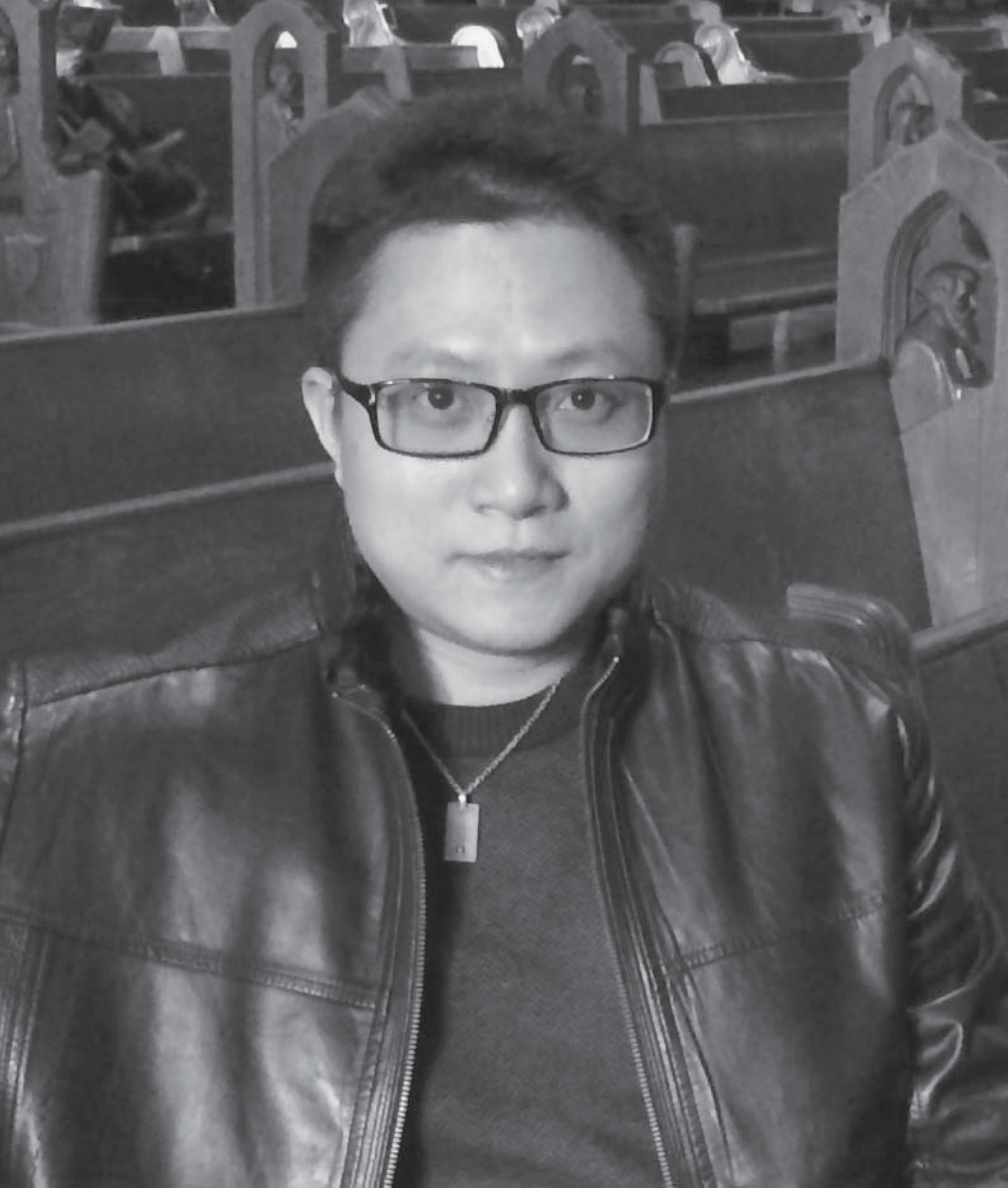}}]{Liang Lin} is CEO of DMAI Great China and a full professor of Computer Science in Sun Yat-sen University. He served as the Executive Director of the SenseTime Group from 2016 to 2018, leading the R\&D teams in developing cutting-edge, deliverable solutions in computer vision, data analysis and mining, and intelligent robotic systems.  He has authored or co-authored more than 200 papers in leading academic journals and conferences (e.g., TPAMI/IJCV, CVPR/ICCV/NIPS/ICML/AAAI). He is an associate editor of IEEE Trans, Human-Machine Systems and IET Computer Vision, and he served as the area/session chair for numerous conferences, such as CVPR, ICME, ICCV, ICMR. He was the recipient of Annual Best Paper Award by Pattern Recognition (Elsevier) in 2018, Dimond Award for best paper in IEEE ICME in 2017, ACM NPAR Best Paper Runners-Up Award in 2010, Google Faculty Award in 2012, award for the best student paper in IEEE ICME in 2014, and Hong Kong Scholars Award in 2014. He is a Fellow of IET.
\end{IEEEbiography}
	
	
	
	
	
	
	

\end{document}